\documentclass[conference, twocolumn]{IEEEtran}
\usepackage{graphicx}
\usepackage{array}
\usepackage{natbib}

\usepackage[utf8]{inputenc}
\usepackage[T1]{fontenc}

\usepackage{listings}
\usepackage{subcaption}
\usepackage[table,xcdraw]{xcolor}
\usepackage{amsmath}
\usepackage{bbm}
\usepackage{pifont}
\usepackage{amssymb}
\definecolor{mydarkblue}{rgb}{0,0.08,0.65}
\usepackage[colorlinks=true,
    linkcolor=mydarkblue,
    citecolor=mydarkblue,
    filecolor=mydarkblue,
    urlcolor=mydarkblue]{hyperref}
\usepackage{cleveref}
\usepackage{soul}
\usepackage[shortlabels]{enumitem}
\usepackage{url}
\usepackage{color}
\usepackage{tikz}
\usepackage{balance}
\usepackage{multirow}
\usepackage{comment}
\usepackage{wrapfig,booktabs}

\usepackage[symbol]{footmisc}
\usepackage{tablefootnote}
\usepackage{tabularx}
\usepackage{placeins}
\usepackage{algorithm}
\usepackage{algpseudocode}
\usepackage{ragged2e}
\usepackage{cuted}
\usepackage{float}

\usepackage{microtype}
\usepackage{xurl}
\DeclareUrlCommand\code{\urlstyle{tt}}

\usepackage{dblfloatfix}

\definecolor{codegreen}{rgb}{0,0.6,0}
\definecolor{codegray}{rgb}{0.5,0.5,0.5}
\definecolor{codepurple}{rgb}{0.58,0,0.82}
\definecolor{backcolour}{rgb}{0.95,0.95,0.92}

\makeatletter
\def\blfootnote{\xdef\@thefnmark{}\@footnotetext}
\makeatother

\lstdefinestyle{mystyle}{
  backgroundcolor=\color{backcolour},   commentstyle=\color{codegreen},
  keywordstyle=\color{magenta},
  numberstyle=\tiny\color{codegray},
  stringstyle=\color{codepurple},
  basicstyle=\ttfamily\footnotesize,
  breakatwhitespace=false,
  breaklines=true,
  captionpos=b,
  keepspaces=true,
  numbers=left,
  numbersep=5pt,
  showspaces=false,
  showstringspaces=false,
  showtabs=false,
  tabsize=2,
}

\AtBeginDocument{%
  \providecommand\BibTeX{{%
    \normalfont B\kern-0.5em{\scshape i\kern-0.25em b}\kern-0.8em\TeX}}}

\IEEEoverridecommandlockouts
\begin{document}

\title{ZUNA1.1: A more flexible EEG foundation model for Denoising and Super-resolution}

% Variable-Length, Any-Montage EEG Reconstruction
% Generalized Channel Infilling for EEG Denoising and Superresolution
% ZUNA1.1: Do Masked Autoencoders Learn Optimal Latent Embeddings for Downstream EEG Tasks?

% \newcommand{\corr}{\textsuperscript{*}}

\author{
Christopher Warner, Jonas Mago, JR Huml, and Beren Millidge
\\[0.5em]
\textbf{Zyphra}
\\
San Francisco, CA \\
%\IEEEauthorblockA{\textsuperscript{1}Zyphra}
% \IEEEauthorblockA{\textsuperscript{*}Corresponding authors: \texttt{chris@zyphra.com}, \texttt{beren@zyphra.com}}
}
\maketitle

\setcounter{page}{1}

\begin{abstract}\normalfont\mdseries
We introduce \texttt{ZUNA1.1}, a 380M-parameter diffusion autoencoder for flexible EEG signal reconstruction. \texttt{ZUNA1.1} is capable of reconstructing variable length sequences of up to 30\,s, with an arbitrary number of EEG channels at arbitrary scalp locations, and can reconstruct arbitrary temporal intervals within channels in addition to reconstructing entire channels. We demonstrate that \texttt{ZUNA1.1} performs at least on par with our earlier \texttt{ZUNA1} model, while being far more flexible and capable of handling a wide range of reconstruction tasks. \texttt{ZUNA1.1} continues to substantially outperform standard EEG denoising and reconstruction methods such as spherical spline interpolation, which is ubiquitously deployed in the MNE package. The \texttt{ZUNA1.1} model is released open source under the permissive Apache 2.0 license. 
\end{abstract}

\begin{figure*}[t]
    \centering
    \includegraphics[width=\linewidth]{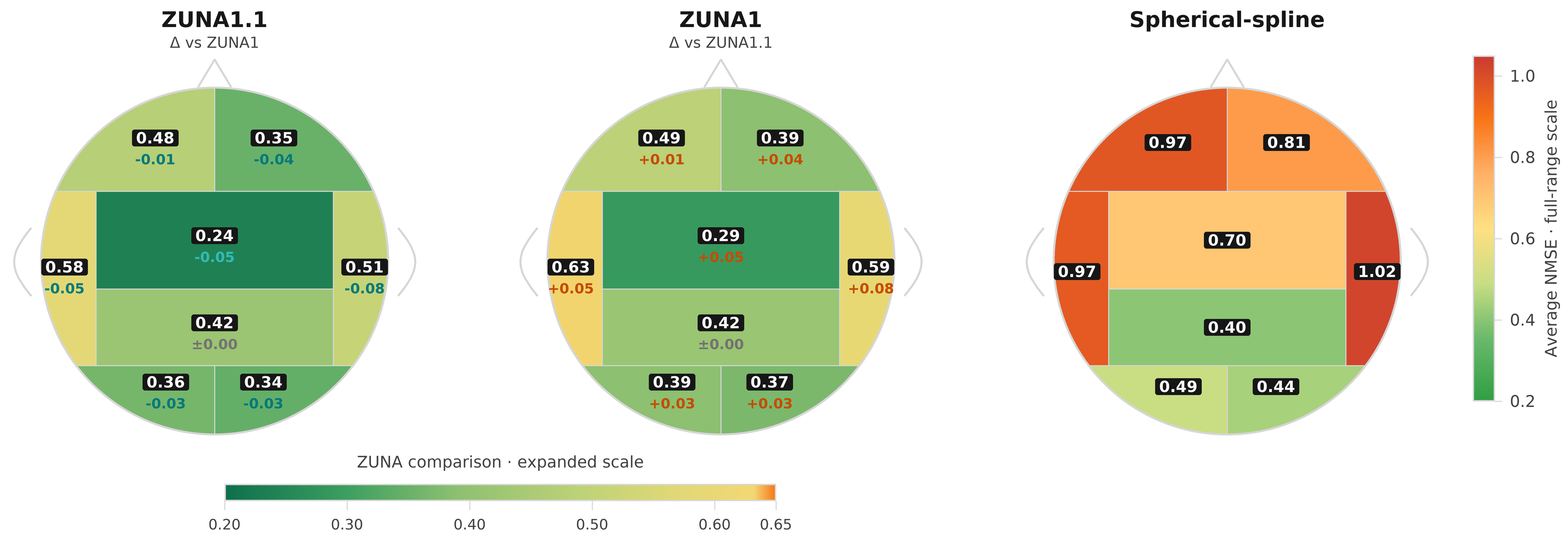}
    \caption{\textbf{Per-region reconstruction accuracy (topographic).} Average NMSE by
    scalp region for \texttt{ZUNA1.1}, \texttt{ZUNA1} (labeled with the $\Delta$
    relative to \texttt{ZUNA1.1}), and spherical-spline, averaged over the four datasets.
    All electrodes in one region are deleted and reconstructed from the remaining seven
    regions. Lower (greener) is better.}
    \label{fig:region_topo}
\end{figure*}

\section{Introduction}
\label{sec:intro}
EEG is the most accessible modality for measuring brain activity, valued in clinical, research, and consumer settings for its simple setup and low cost. Although the skull blurs the signal at each electrode to approximately an average of over a billion neurons, scalp recordings still carry substantial decodable information. Prior work has classified sleep stages \citep{supratak2017deepsleepnet, li2022deep}, decoded emotional states \citep{liang2019unsupervised, duan2020machine}, and tracked attentional focus \citep{su2022stanet, kaushik2022decoding}. This success has driven adoption beyond the lab. A growing range of consumer-grade headsets now makes EEG affordable and easy to deploy outside research settings. However, consumer and inexpensive lab-grade headsets often have a much smaller number of channels and poorer data quality than full lab-grade devices. Moreover, all EEG data collection methods often suffer from occasional corruption, artifacts, and general low quality recordings. For instance, electrodes can lose contact or their signal can weaken, motion artifacts can disrupt entire groups of channels at a time, and general noise remains prevalent.

To address these limitations, we recently introduced \texttt{ZUNA1} \citep{warner2026zuna}, a 380M-parameter masked diffusion autoencoder that repairs bad channels and upsamples consumer-grade recordings to as many as 256 channels, matching the highest quality lab-grade systems. \texttt{ZUNA1} tokenizes multichannel EEG into short temporal windows and injects spatiotemporal structure via a 4D rotary positional encoding over (x,y,z,t), enabling inference on arbitrary channel subsets and positions. It substantially outperformed spherical-spline interpolation \citep{perrin1989sphericalspline} for channel reconstruction across heterogeneous datasets. Our thinking was that much of the EEG signal lost due to various sources of noise was contained redundantly across the channel montage as a whole, however, it is encoded nonlinearly and is thus inaccessible to conventional spline interpolation methods. Instead, by training EEG foundation models, which can learn generalizable patterns across millions of channel-hours of EEG data, these foundation models would be able to provide a similar boost to that which foundation models have produced in other modalities such as text, images, and audio. 

In addition to ZUNA1, a number of other EEG foundation models have been developed in recent years \citep{kuruppu2025eeg}, differing mainly in their self-supervised objective. Contrastive models such as BENDR \citep{kostas2021bendr} and SleepFM \citep{thapa2024sleepfm}, and a large family of masked-reconstruction models including BIOT \citep{yang2023biot}, LaBraM \citep{jiang2024large}, CBraMod \citep{wang2024cbramod}, Brant \citep{zhang2023brant}, BrainWave \citep{yuan2024brainwave}, LUNA \citep{doner2025luna}, and REVE \citep{elouahidi2025reve}, dominate, alongside autoregressive \citep{jiang2024neurolm, cui2024neuro} and latent-predictive (JEPA) \citep{wang2024eegpt, panchavati2026laya} approaches. \texttt{ZUNA} differs from this body of work in three respects. First, unlike all but a handful of recent efforts, it is trained at scale (${\sim}3.5$M channel-hours). Second, it uses a position-conditioned \emph{diffusion} architecture rather than a discriminative encoder. Third, and most importantly, nearly all of these models are optimized to learn a general-purpose latent space which can then be fine-tuned or linearly probed for downstream \emph{classification}, whereas \texttt{ZUNA}'s training objective is the signal \emph{reconstruction} itself. This makes \texttt{ZUNA} directly suited to flexibly infilling missing or corrupted EEG, the problem we target, rather than producing features that a separate model must then decode.

However, deploying ZUNA1 to real workflows revealed two practical limitations. Firstly, the model operated only on fixed 5-second windows, with no support for longer or shorter sub-segments. This is clearly a major limitation since almost all naturally occurring EEG data is not exactly five seconds in duration, and chopping existing data into the 5-second chunks often causes unnatural boundary artifacts.   Secondly, the original ZUNA1 training used a single dropout scheme that randomly sub-selected channels. The problem of the random dropout scheme was that it does not necessarily match the patterns of missing channels in real world cases. In real-world recordings, an individual channel completely dropping and remaining unavailable for the entire duration is rare. A much more common pattern is bursts of errors, often individual or correlated across many channels due to temporary electrode glitches or motion artifacts. Moreover, consumer headsets impose specific sparse layouts of channels that tend to fail in spatially or temporally correlated bursts.

To address these limitations, we introduce \texttt{ZUNA1.1}, which retains the architecture of \texttt{ZUNA1} and addresses each of the limitations above through the following contributions:

\begin{itemize}
    \item \emph{Variable-length training}: \texttt{ZUNA1.1} trains on randomly cropped windows from $0.5$ to $30$ seconds, packed into batches via flex attention with sample-aware masks for efficient training and inference.
    \item \emph{Implicit augmentation}: \texttt{ZUNA1.1} sees each underlying signal under multiple transformations across exposures: random temporal crops combined with per-window z-score normalization, and two pre-computed filter variants of all data (highpass+notch and bandpass). This provides substantial implicit data augmentation that lets us train for many more passes over the corpus without overfitting.
    \item \emph{Quality-aware preprocessing}: each recording is paired with a per-channel, per-second quality matrix computed from amplitude extremes, segment-level standard deviation, and flatness/clipping detectors, with thresholds applied at load time rather than at preprocessing time, so the same on-disk corpus can be re-thresholded for any desired stringency without re-preprocessing.
    \item \emph{Diverse dropout mixture}: training uses a mixture of eight different dropout schemes spanning uncorrelated, contiguous, anatomical, montage-density, and consumer-headset patterns, broadening the range of realistic infilling scenarios the model handles without retraining.
    \item \emph{Expanded corpus and training scale}: quality-aware loading and variable-length training together let us recover signal from partially noisy channels and ingest pre-epoched datasets the original pipeline could not use, growing the training corpus from approximately 2M channel-hours \citep{warner2026zuna} to approximately 3.5M channel-hours. In addition to the increased training corpus, ZUNA1.1 was trained for substantially more steps.  
\end{itemize}

As with \texttt{ZUNA1}, we release the \texttt{ZUNA1.1} weights under the Apache License 2.0 on Hugging Face, together with inference and preprocessing code on GitHub. The package is available via \texttt{pip install zuna}.

\begin{figure*}[t]
    \centering
    \includegraphics[width=\linewidth]{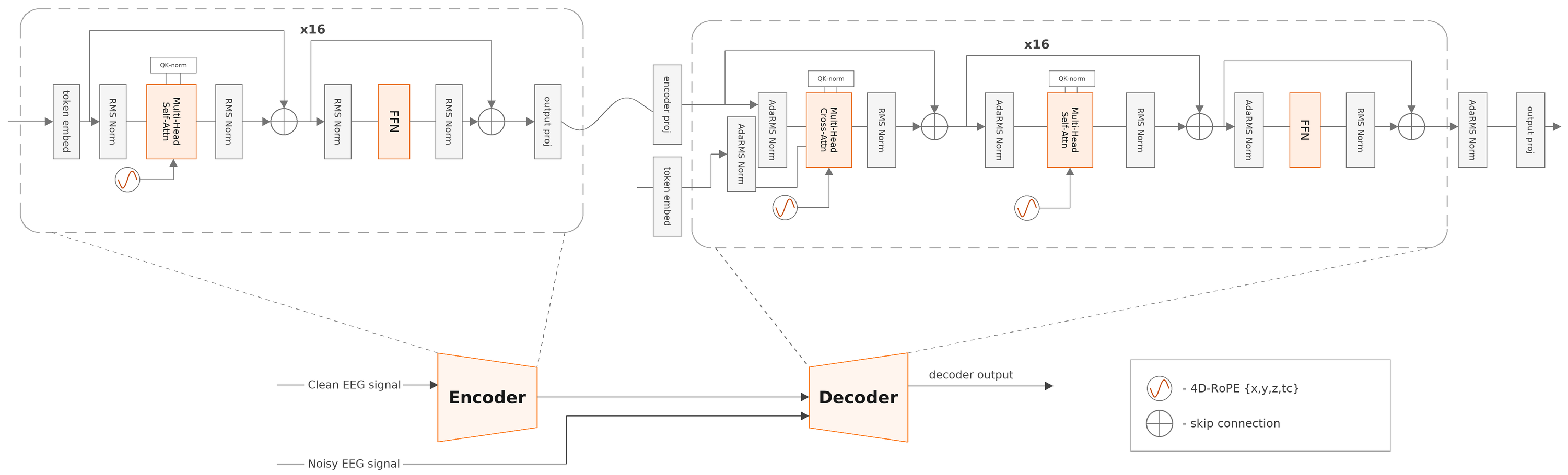}
    \caption{\textbf{\texttt{ZUNA1.1} architecture.} A transformer-based
    diffusion autoencoder. The main modifications to the \texttt{ZUNA1} architecture are adding sandwich norm and QK-norm to improve
    training stability. This allowed us to remove Adaptive Loss Weighting from the decoder RF loss term.}
    \label{fig:architecture}
\end{figure*}

\section{Background and Architecture}
\label{sec:background}

\texttt{ZUNA1.1} is built on a slightly modified version of the ZUNA1 model architecture and we refer readers to our original release \citep{warner2026zuna} for a full overview of the model specifications. \texttt{ZUNA1.1} is a transformer-based encoder--decoder \emph{diffusion autoencoder} for masked channel reconstruction. The encoder produces latent representations that are injected into the decoder. The decoder is trained with a rectified-flow loss. The encoder latent is regularized with an auxiliary MMD loss. In an effort to improve the quality of latent representations for downstream tasks, we experimented with reducing the weight on the MMD regularization term and discuss these experiments further below. To improve training stability, we added sandwich normalization \citep{ding2021sandwichnorm} and Query-Key normalization \citep{henry2020qknorm} to each layer in the encoder and decoder of the \texttt{ZUNA1} architecture, shown in Figure \ref{fig:architecture}. These architectural improvements allowed us to remove adaptive loss weighting \citep{geng2024alw} from the decoder RF-loss without reintroducing training instabilities.

\texttt{ZUNA1.1} retains the same discretized positional encoding scheme as \texttt{ZUNA1} implemented with 4D-RoPE but  extends training support to segment lengths from $0.5$ to $30$ seconds. This encoding scheme allows the model to process arbitrary sequence lengths (up to 30\,s) as well as an arbitrary number of channels at arbitrary scalp locations. The total parameter count is unchanged at 380M. In early ablations, we explored scaling the model size to 1.2B parameters, and found that scaling up did not meaningfully improve reconstruction or performance on downstream tasks.

\section{Method}
\label{sec:method}

\begin{table*}[t]
\centering
\setlength{\tabcolsep}{4pt}
\caption{\textbf{Channel-dropout schemes and their sampling weights across the three training stages.} Stage 1 ($0$--$250$k steps) samples all eight schemes with the $1/8$ weights shown; Stage 2 ($250$--$440$k steps) samples all eight schemes with the (non-uniform) weights shown; Stage 3 ($440$--$580$k steps) drops the four \emph{layout} schemes (bottom four, whose loss had plateaued) and samples the four \emph{structure} schemes (top four) uniformly. Within any selected scheme, dropout is applied with probability $p_\text{drop}=0.9$.}
\label{tab:dropout_schemes}
\begin{tabularx}{\textwidth}{l l c c c X}
\toprule
Scheme & Pattern & Stage 1 & Stage 2 & Stage 3 & Description \\
\midrule
\texttt{full-channel-random} & spatial, uncorrelated & $0.125$ & $0.275$ & $0.25$ & \texttt{ZUNA1}'s original scheme: drops a random subset of entire channels for the entire sample window. \\
\texttt{random-uniform} & unstructured & $0.125$ & $0.075$ & $0.25$ & Drops individual (channel, coarse-time) tokens independently. \\
\texttt{full-time-pt-random} & temporal & $0.125$ & $0.125$ & $0.25$ & Drops disjoint contiguous blocks of coarse-time steps across all channels ($\in[0.25,2]$s), modeling brief whole-recording artifacts. \\
\texttt{correlated-channel-time} & spatio-temporal & $0.125$ & $0.125$ & $0.25$ & As above, but each dropped time block is removed from a random subset of channels: spatially and temporally clustered dropout. \\
\midrule
\texttt{standard-montage} & montage density & $0.125$ & $0.125$ & --- & Keeps only channels nearest to a standard 8-, 16-, 32-, or 64-channel 10--20 montage. \\
\texttt{random-montage} & montage density & $0.125$ & $0.125$ & --- & Greedily removes one channel of the closest remaining electrode pair down to a random count (8/16/32/64): sparse, uniform coverage. \\
\texttt{brain-region} & anatomical & $0.125$ & $0.125$ & --- & Partitions channels into eight scalp regions by 3D coordinates and drops all channels in a random subset of regions. \\
\texttt{consumer-eeg} & device layout & $0.125$ & $0.025$ & --- & Keeps only channels nearest the electrode positions of a randomly selected consumer headset: Muse (4), Emotiv Insight (5), Dreem (5), Neurosity Crown (8), Unicorn (8), OpenBCI Cyton (8), Emotiv EPOC (14), Flex32 (32). \\
\bottomrule
\end{tabularx}
\end{table*}

\subsection{Variable-length sequence training}
\label{sec:method:varlen}

In \texttt{ZUNA1}, every training sample was a fixed five-second segment. \texttt{ZUNA1.1} instead samples a window length $w$ uniformly at random per training example, snapped to a multiple of the $\tau = 0.125\,\text{s}$ coarse-time unit. Window lengths are drawn from a piecewise-uniform distribution over four bins:
\begin{center}
\begin{tabular}{lcc}
\toprule
Bin & Range & Sampling weight \\
\midrule
Very short  & $0.5$--$1.0$\,s     & $0.20$ \\
Short       & $1.0$--$5.0$\,s     & $0.30$ \\
Medium      & $5.0$--$10.0$\,s    & $0.30$ \\
Long        & $10.0$--$30.0$\,s   & $0.20$ \\
\bottomrule
\end{tabular}
\end{center}

This distribution focuses on the $1$--$10$\,s range, which we found to be the most common operating regime in downstream applications, while preserving meaningful coverage of both very short windows (which exercise the model's short-context behavior) and very long windows (which let the model learn long-range structure).

A consequence of variable-length training is that the per-sample token count varies dramatically. We extend the sample-packing strategy from \texttt{ZUNA1} to handle this efficiently. Random-length windows are packed contiguously into a batch and padded to fit a target packed sequence length. Flex attention is then used to compute a sample-aware attention mask that prevents tokens from one sample attending to tokens of another. This approach allows a compiled model to efficiently process a large batch of tokens from multiple samples in parallel, filling GPU memory and maximizing utilization.

\begin{figure*}[t]
    \centering
    \includegraphics[width=\linewidth]{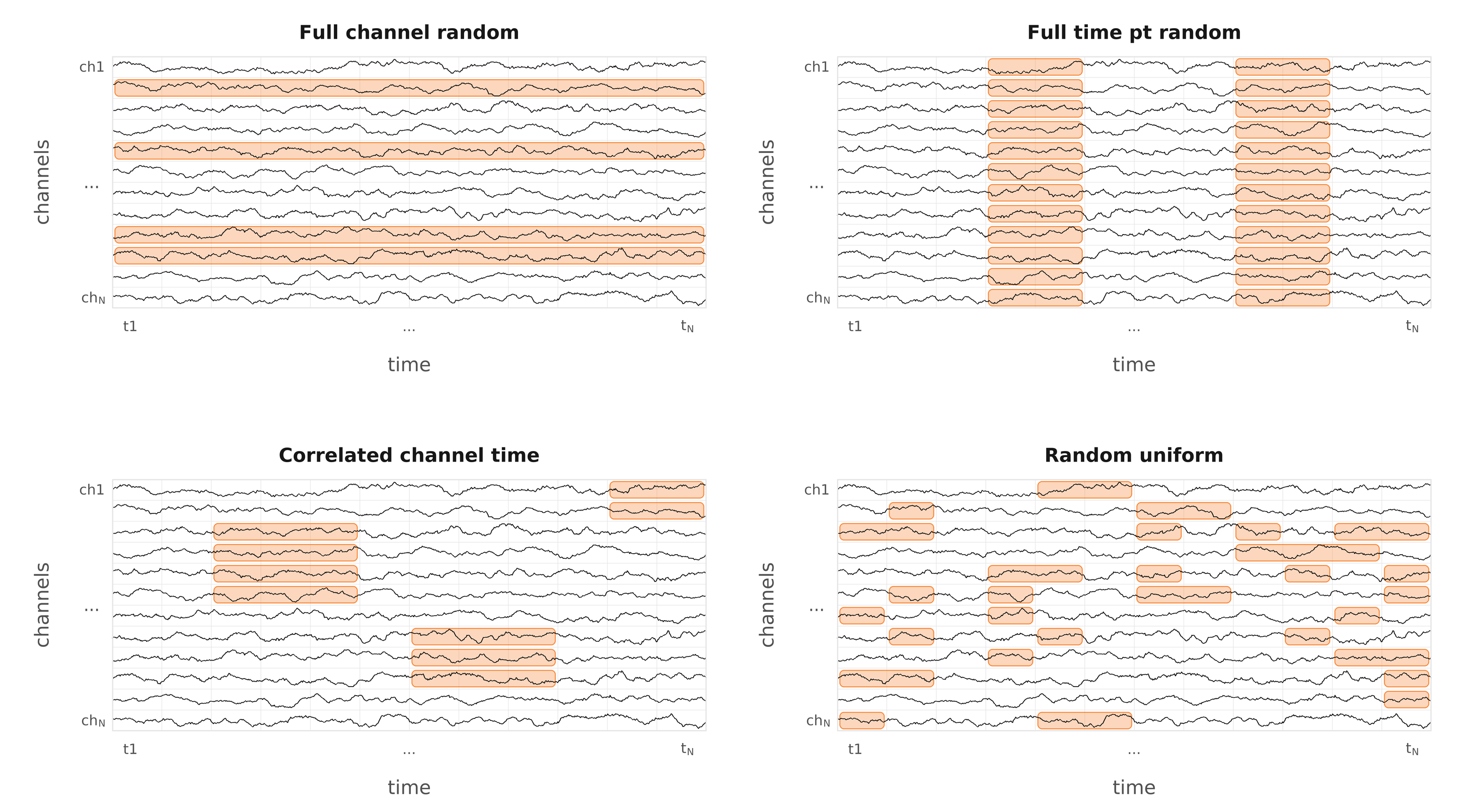}
    \caption{\textbf{The four channel-dropout schemes retained throughout training.}
    Each panel shows one training sample: channels run along the vertical axis and
    coarse-time windows ($\tau = 0.125$\,s) along the horizontal axis. Observed tokens (unshaded traces) are passed to the encoder; the
    shaded spans are set to zero in the encoder and reconstructed by the decoder based on observed signals.
    \texttt{full-channel-random} removes entire channels;
    \texttt{full-time-pt-random} removes short time spans across all
    channels; \texttt{correlated-channel-time} removes those spans
    from a subset of channels; \texttt{random-uniform} scatters
    missing tokens independently. These are the four \emph{structural} dropout
    schemes retained for the full training run (\cref{sec:method:dropout}); the four
    \emph{layout} schemes, phased out after $440$k steps, are not shown.}
\label{fig:dropout_schematic}
\end{figure*}

\subsection{Quality-aware preprocessing}
\label{sec:method:quality}

Previously, \texttt{ZUNA1}'s preprocessing pipeline operated at the recording level such that a channel which exceeded noise thresholds for any portion of a recording was zeroed for the whole recording, and epochs with too many bad channels were discarded entirely. Instead, \texttt{ZUNA1.1}'s pipeline stores the full continuous recording on disk and computes a per-channel, per-second \emph{quality matrix} $Q \in [0, 1]^{C \times S}$, where $C$ is the number of channels and $S$ is the number of one-second segments in the recording. The quality score combines three markers, each comparing a segment against the statistics of its own channel (or, for flatness $q^{\text{flat}}$, of the whole recording). Let $\sigma_{c,s}$ denote the standard deviation of channel $c$ in segment $s$, $\tilde{\sigma}_c$ the median of $\sigma_{c,s}$ over segments, and $\rho_{c,s}$ the segment's peak-to-peak amplitude relative to its channel-median peak-to-peak. The three scores are
\begin{align*}
q^{\text{var}}_{c,s} &= \exp\!\left(-\tfrac{1}{2}\left(\tfrac{\log(\sigma_{c,s}/\tilde{\sigma}_c)}{0.7}\right)^{\!2}\right), \\
q^{\text{ptp}}_{c,s} &= \left(1 + e^{\,2(\rho_{c,s} - 3.5)}\right)^{-1}, \\
q^{\text{flat}}_{c,s} &= \mathbbm{1}\!\left[\sigma_{c,s} \geq 0.05 \cdot \operatorname{median}_c \tilde{\sigma}_c\right],
\end{align*}
and the final quality is their element-wise minimum, $Q_{c,s} = \min(q^{\text{var}}_{c,s},\, q^{\text{ptp}}_{c,s},\, q^{\text{flat}}_{c,s})$. The variance score penalizes both noise bursts and flat stretches (deviations of the segment's variance from the channel's typical variance, measured in log-space); the peak-to-peak score falls off steeply for large transient artifacts; and the flatness score hard-zeroes dead channels, including reference channels that are flat by construction.

The quality matrix is saved alongside the recording as a separate mmap file, and quality thresholds are applied at \emph{load time}. The data loader takes two thresholds, \code{min_quality_any} and \code{min_quality_mean}: a channel is included in the current training window if every one-second segment overlapping the window has quality $\geq$ \code{min_quality_any} \emph{and} the mean quality across those segments is $\geq$ \code{min_quality_mean} (set to $0.1$ and $0.3$ respectively for the released model). This means that a channel which is noisy for part of a recording can still contribute usable training data drawn from the clean portions, and the same on-disk corpus can be re-thresholded for any desired quality stringency without rerunning preprocessing.

The recording is stored as a memory-mapped \texttt{float32} array of shape $(C, T)$, where $T$ is the number of samples in the recording at $256\,\text{Hz}$. Pre-epoched recordings (which the original \texttt{ZUNA1} pipeline could not ingest) are stored as $(E, C, T_e)$ where $E$ is the number of epochs and $T_e$ is the number of samples per epoch. Per-channel z-score normalization parameters used at preprocessing time are stored in the recording's metadata so that the operation can be reverted if needed.

\subsection{Multiple filter variants}
\label{sec:method:filters}

EEG analyses use a wide variety of upstream filtering conventions, and there is no universally agreed default. To allow \texttt{ZUNA1.1} to interoperate with users' existing pipelines, we precompute and save two filter variants of every recording:

\begin{itemize}
    \item \textbf{Notch variant}: A $0.01\,\text{Hz}$ highpass plus zero-phase notch filters at powerline frequencies and their harmonics ($50$, $60$, $100$, $120$, $150$, $180$, $200$, and $240\,\text{Hz}$, restricted to frequencies below Nyquist). Since our corpus mixes recording sites on $50\,\text{Hz}$ and $60\,\text{Hz}$ mains power, both families are removed. This variant preserves the broadband signal, including slow drifts and high-frequency content, apart from line noise.
    \item \textbf{Bandpass variant}: A $0.1$--$45\,\text{Hz}$ bandpass filter applied independently to the raw recording. Suitable for users whose downstream analyses operate in the conventional clinical EEG band.
\end{itemize}

Both variants are sampled with equal probability during training: each exposure of a recording samples one of the two, which contributes to the implicit augmentation described in \cref{sec:method:varlen}. Users can select a variant at inference time to match their preferred preprocessing convention. Filtering is applied with EEGLAB-compatible boundary handling (\texttt{skip\_by\_annotation}) so that boundary annotations from re-concatenated cleaned epochs do not introduce filter ringing across discontinuities.

\subsection{Diverse channel dropout mixture}
\label{sec:method:dropout}

Perhaps the most consequential advancement is the dropout mixture used at training time. The original \texttt{ZUNA1} model was trained exclusively with a single dropout scheme that randomly removed whole channels with uniform probability. However, this did not match many of the patterns of degradation or noise encountered in practice. For instance, consumer EEG devices have very specific sparse channel layouts, motion artifacts tend to corrupt anatomically clustered channels rather than randomly distributed channels, and electrode dropout is often correlated in time as well as in space.

\begin{figure*}[t]
    \centering
    \includegraphics[width=0.8\textwidth]{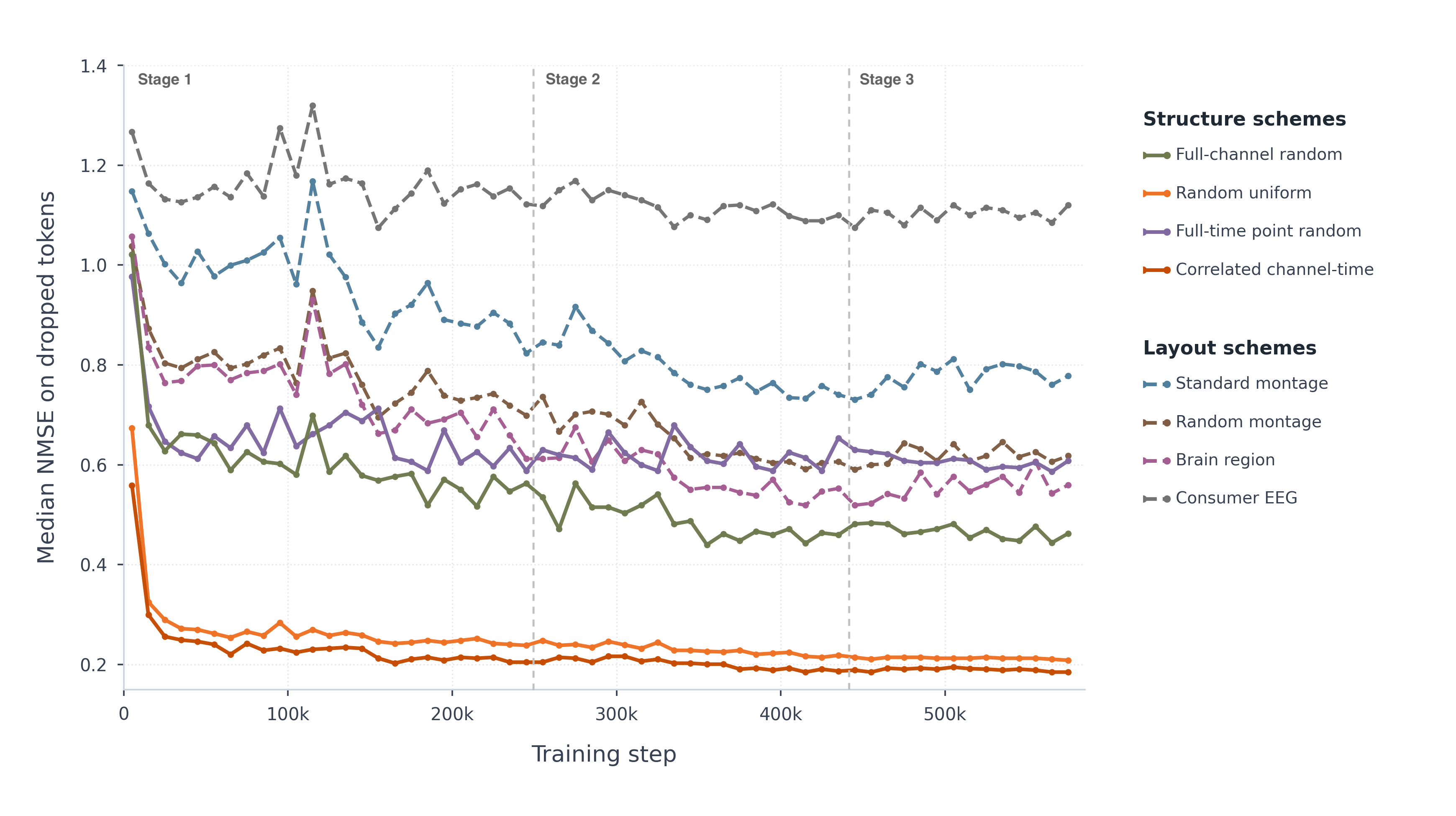}
    \caption{\textbf{Reconstruction performance for each dropout scheme throughout training.} Median NMSE (lower is better) on the frozen evaluation dataset for each dropout scheme. See Table~\ref{tab:dropout_schemes} for details on how the dropout schemes were adjusted during training.}
    \label{fig:NMSE_vs_train_step_dropout}
\end{figure*}

For \texttt{ZUNA1.1}, we instead trained on a diverse mixture of eight dropout schemes, summarized in Table~\ref{tab:dropout_schemes}, which fall into two families: four that control the dropout \emph{structure} and four that control the \emph{layout} of the retained channels. During training, we used a three-stage curriculum. See Table~\ref{tab:dropout_schemes} for details. For the first $250$k steps we sample all eight schemes using uniform weights. From $250$k - $440$k steps, observing that evaluation metrics had plateaued for many of the dropout schemes (cf. Fig. \ref{fig:NMSE_vs_train_step_dropout}) because they were too hard or already learned, we changed the parameters of individual dropout schemes to adjust their difficulty level and adjusted the weight of each scheme as reflected in Stage 2. This included increasing overall dropout probability from $0.9$ to $0.99$ because the model had perfectly learned the easy task of reconstructing non-dropped tokens. Finally, in Stage 3, from $440$k - $580$k steps, we retained only the 4 structure dropout schemes at uniform weighting. %Within any selected scheme, token dropout is applied to a sample with probability $p_\text{drop} = 0.9$, matching the \texttt{ZUNA1} dropout rate. 
We found the four layout schemes were substantially more challenging for the model to learn than the structural schemes. For instance, the consumer-headset scheme can require reconstructing a full 256-channel montage from as few as four electrodes.

The two families of dropout schemes play complementary roles. The four \emph{structure} schemes control the \emph{pattern} of dropout, from fully unstructured token dropout through to spatio-temporally correlated dropout that mimics motion artifacts and transient electrode failures. The four \emph{layout} schemes instead control which channels are retained. The montage and consumer-headset schemes select channels by matching each target electrode position to the nearest recorded electrode in 3D space, so they apply to any input montage, while the brain-region scheme partitions the recorded electrodes themselves by scalp coordinates. Together, the eight schemes expose the model to a wide range of infilling scenarios we target at deployment: upsampling a sparse consumer headset to a dense research montage, repairing an anatomically-clustered block of bad channels, and filling temporal gaps from transmission dropouts. Because the layout schemes proved hardest and appeared to quickly saturate at a high loss, we concentrated the final training stretch on the four structure schemes which were amenable to further progress.  However, since we trained on the layout schemes for the majority ($440$k) of its steps, \texttt{ZUNA1.1} retains strong montage-upsampling and region-reconstruction performance at deployment.

It should be noted that by combining the random crop dropout schemes and the per-window z-score normalization, we achieve substantial \emph{implicit data augmentation}. Each time the same recording is loaded, a different temporal crop is randomly sampled, and the per-channel mean and standard deviation used for normalization are computed locally on that crop. The resulting numerical input therefore differs across exposures even when the underlying signal is identical, which broadens the input distribution the model sees and reduces effective overfitting. This effect is essentially free, requiring no explicit augmentation pipeline. Supporting this, we trained for 6 epochs without noticing any signs of overfitting in validation loss.

\begin{figure*}[t]
    \centering
    \includegraphics[width=0.8\linewidth]{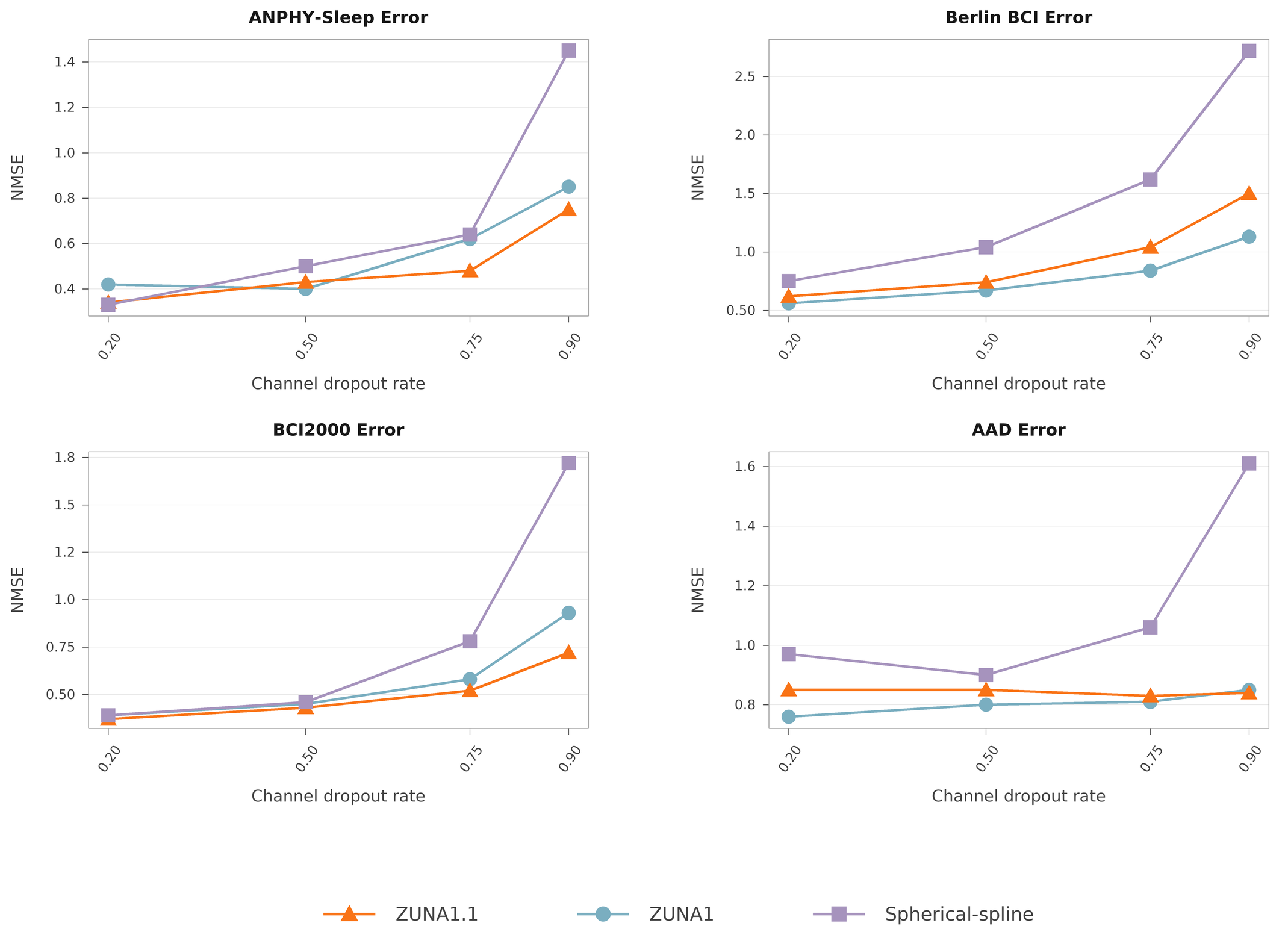}
    \caption{\textbf{Reconstruction accuracy as channels drop.} Reconstruction NMSE
    versus channel-dropout rate on four held-out datasets (ANPHY-Sleep, BerlinBCI,
    BCI2000, AAD), comparing \texttt{ZUNA1.1}, \texttt{ZUNA1}, and MNE spherical-spline
    interpolation; lower is better. \texttt{ZUNA1.1} matches or improves on
    \texttt{ZUNA1}, and both clearly outperform spline interpolation, with the gap
    widening as more channels are removed. Evaluation restricted to $5$\,s samples for
    comparability with \texttt{ZUNA1}.}
    \label{fig:dropout_curves}
\end{figure*}

\begin{figure*}[t]
    \centering
    \includegraphics[width=0.8\linewidth]{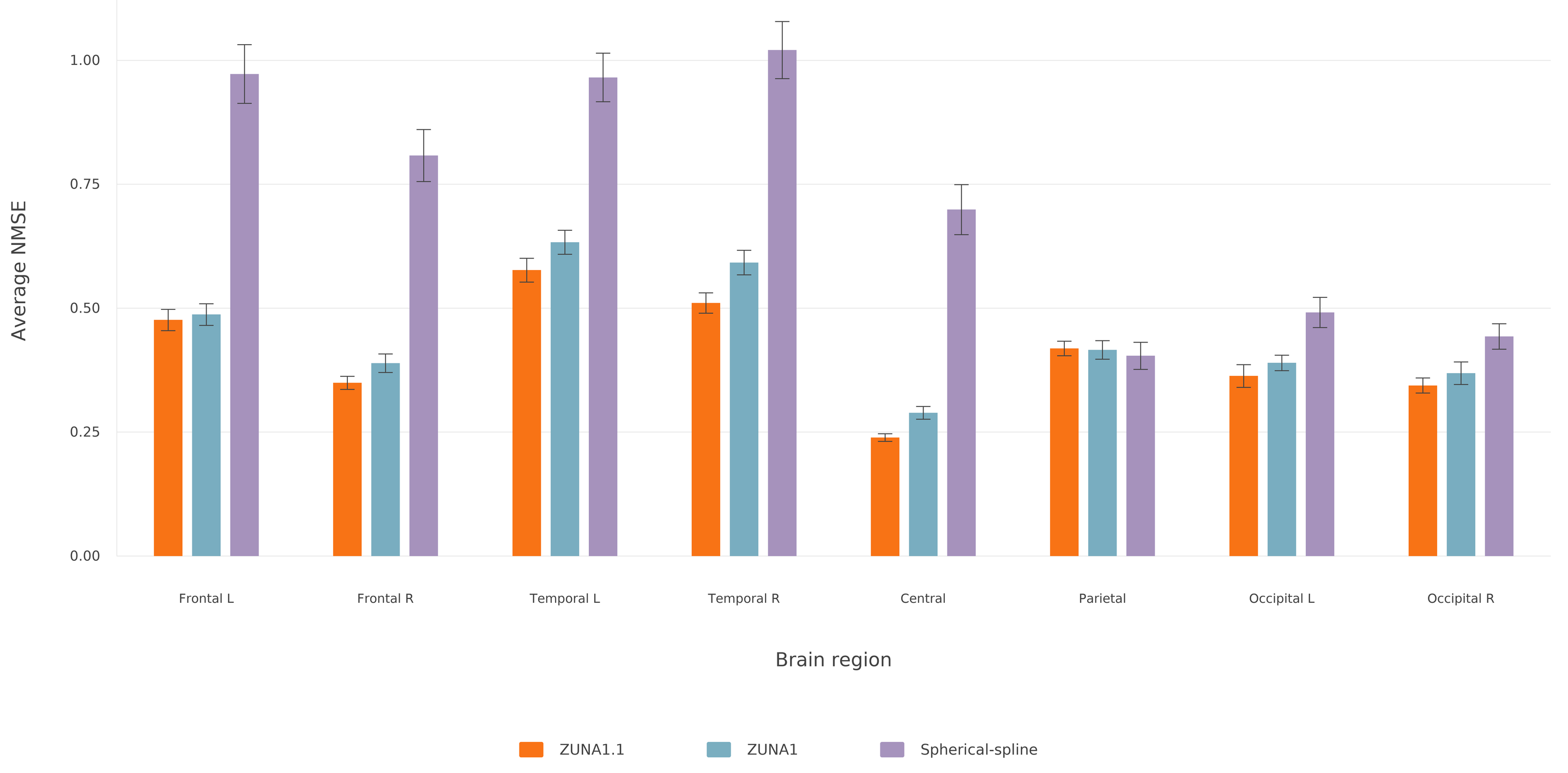}
    \caption{\textbf{Per-region reconstruction error.} The region-occlusion evaluation
    of \cref{fig:region_topo} shown as grouped bars: average NMSE across the four
    datasets with error bars giving the propagated standard deviation. Lower is better.}
    \label{fig:region_bars}
\end{figure*}

\subsection{Training details}
\label{sec:method:training}

\texttt{ZUNA1.1} was trained for 580k steps on approximately 3.5 million channel-hours of processed and cleaned EEG data. Since predicting the non-dropped tokens is a fairly trivial identity operation, which the model learns extremely quickly, we downweight the loss contribution of the non-dropped tokens to 10\% of the weight for dropped tokens, concentrating the objective on the infilling task while still lightly supervising the retained signal.. We used the standard weight decay of 0.1 on all layers except for embedding and norm layers. 
%The rectified-flow decoder loss weights the reconstruction of dropped (masked) tokens by $1.0$ and of kept (observed) tokens by $0.1$, concentrating the objective on the infilling task while still lightly supervising the retained signal.

\texttt{ZUNA1.1} is trained with AdamW ($\beta_1 = 0.9$, $\beta_2 = 0.95$) with a weight decay of $0.1$ applied to the linear, MLP, and attention-projection weights but not to the embedding or normalization parameters. The learning rate is cosine-annealed from a peak of $5\times10^{-4}$ to $9\times10^{-5}$ over $580$k steps, following a $1{,}000$-step linear warmup. The model trains in \texttt{bfloat16} with normalization layers and the residual stream kept in \texttt{float32} for stability, and is compiled end-to-end with \texttt{torch.compile}. We additionally maintain an exponential moving average (EMA) version of the model weights with decay $0.9999$ per step, which is used for evaluation and inference. We ran data-parallel training across 6 nodes with a per-rank micro-batch of 22,000 packed tokens and 2 gradient-accumulation steps; giving a global batch size of roughly $2$M tokens per optimizer step. %The rectified-flow decoder loss weights the reconstruction of dropped (masked) tokens by 1.0 and of kept (observed) tokens by 0.1, concentrating the objective on the infilling task while still lightly supervising the retained signal. 

Training at this scale on heterogeneous public EEG surfaced stability issues worth recording. Rare extreme samples produced sudden gradient-norm spikes followed by loss excursions, and in a handful of cases NaN losses. We addressed these with three safeguards. First, optimizer updates are skipped entirely whenever a non-finite loss is detected. Secondly, we add an epsilon to the z-score denominator to protect against zero-variance (flat or clipped) channel segments. Thirdly, we conducted a corpus scan and removed a small set of recordings whose data and quality matrices contained NaNs. The removed data comprised approximately $0.3\%$ of our total dataset, and we traced the nans to conversion artifacts in three sleep datasets. With architectural improvements and these mitigations in place, we were able to train \texttt{ZUNA1.1} for $4\times$ as many steps as \texttt{ZUNA1} with no signs of instability. 

\section{Results}
\label{sec:results}

\subsection{Reconstruction}

To compare to previous work in \citep{warner2026zuna}, we evaluate \texttt{ZUNA1.1} on four carefully chosen datasets that represent a diverse set of tasks and channel counts, ranging from sleep to motor control, each with 5 second samples that \texttt{ZUNA1} can operate on. Note that 5 second samples are also at the heart of \texttt{ZUNA1.1}'s training distribution. The first experiment repeats Figure 4 of the original \texttt{ZUNA1} paper, where we test the sensitivity of the reconstruction methods with respect to the proportion of deleted channels, progressively deleting more channels and giving the models less information to reconstruct channels. In this experiment, deleted channels are chosen uniformly at random, and are deleted for all time in the sample (matching the dropout training scheme of \texttt{ZUNA1}). We find in Figure \ref{fig:dropout_curves} that, on average, the performance difference between \texttt{ZUNA1.1} and \texttt{ZUNA1} is relatively small, where \texttt{ZUNA1.1} outperforms \texttt{ZUNA1}, on average but not uniformly across all datasets. However, we unequivocally find that both methods are substantially better than spherical-spline interpolation. This performance comes in conjunction with \texttt{ZUNA1.1} being trained on a wider range of sample durations with more diverse dropout schemes, which represent more realistic experimental setups such that \texttt{ZUNA1.1} has much greater range and usability for real-world EEG signals.

Since spherical spline interpolation heavily leverages proximal channels to predict missing signals, it can maintain reasonable performance predicting channels when signal is retained in nearby channels. To stress test the channel prediction methods, we stratified channels into eight regions and dropped all channels from each region, predicting them from the remaining seven regions. Figures \ref{fig:region_bars} and \ref{fig:region_topo} illustrate that \texttt{ZUNA1.1} performs at least as well as \texttt{ZUNA1} and both outperform the spherical spline method on this reconstruction task.

\begin{figure*}[t]
    \centering
    \begin{subfigure}[t]{0.49\linewidth}
        \centering
        \includegraphics[width=\linewidth]{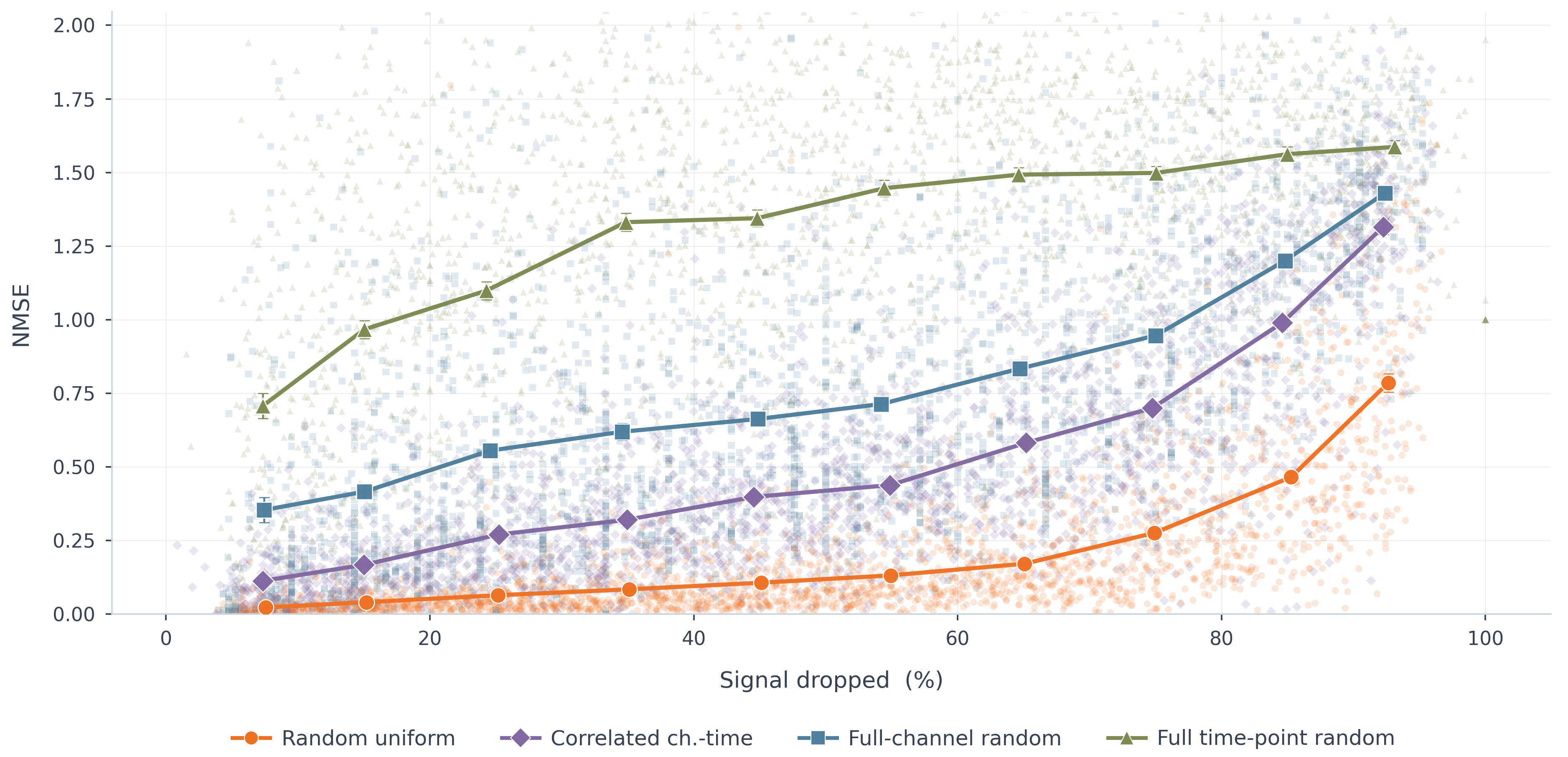}
        \caption{NMSE vs token dropout percentage.}
        \label{fig:nmse_vs_pct_dropout}
    \end{subfigure}
    \hfill
    \begin{subfigure}[t]{0.49\linewidth}
        \centering
        \includegraphics[width=\linewidth]{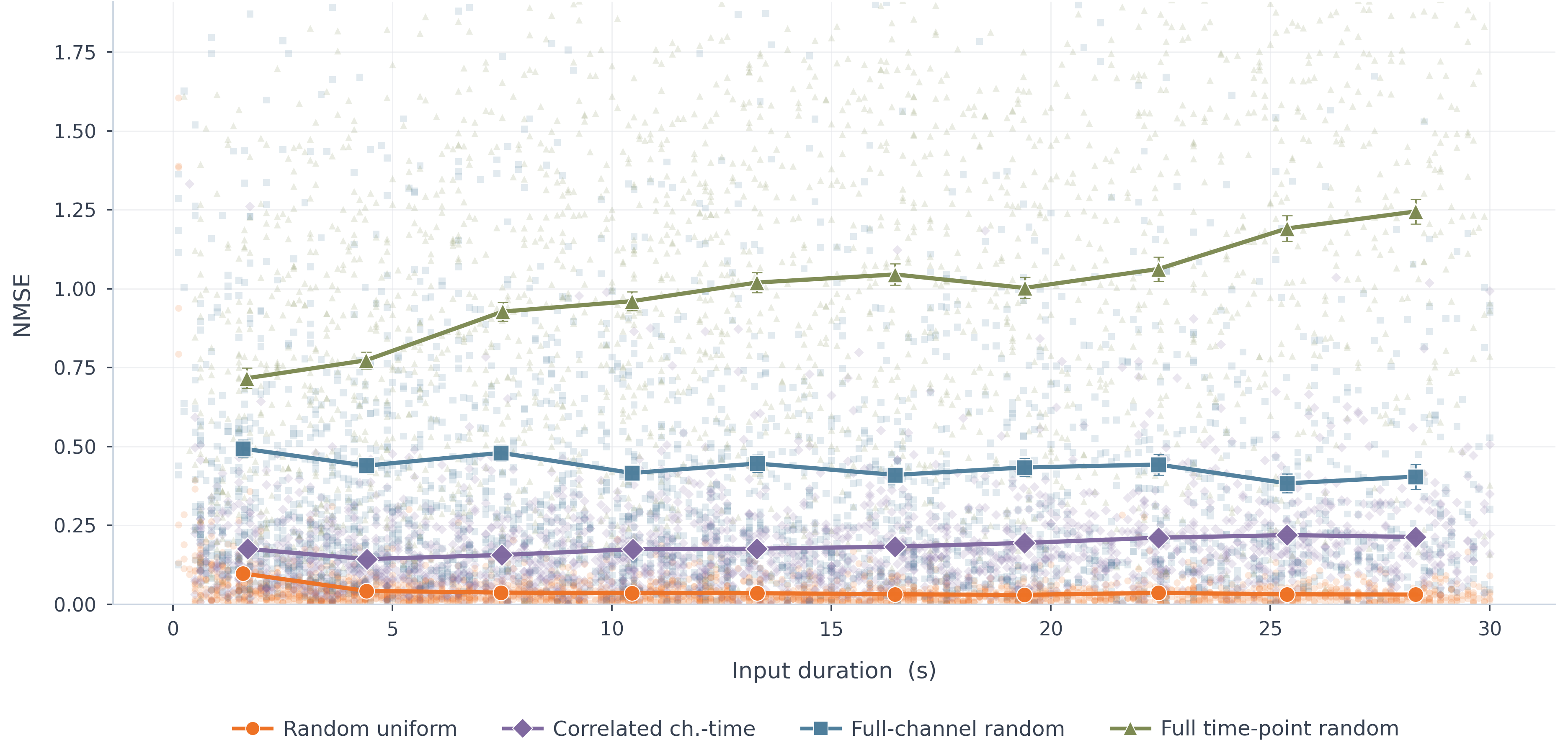}
        \caption{NMSE vs sample duration with token dropout fixed at $15\%$.}
        \label{fig:nmse_vs_duration}
    \end{subfigure}
    \caption{\textbf{Reconstruction performance for 4 structural dropout schemes} on masked tokens in Evaluation dataset. NMSE (lower is better). Description in Figure \ref{fig:dropout_schematic}}
    \label{fig:nmse_v_4dropouts}
\end{figure*}

As discussed above, two major areas of advancement for \texttt{ZUNA1.1} relative to \texttt{ZUNA1} are the ability to process more varied sample durations (from 0.5 - 30 seconds) and the ability to predict missing data from an assortment of realistic structural dropout patterns. Figure \ref{fig:nmse_v_4dropouts} illustrates the model's reconstruction accuracy for the four different structural dropout schemes retained throughout training for an evaluation dataset as we vary what percentage of tokens were masked in panel a, and the duration of the sample with a fixed $15\%$ of tokens masked in panel b.

\begin{figure*}[htbp]
    \centering
    \begin{subfigure}[b]{0.32\textwidth}
        \centering
        \includegraphics[width=\textwidth]{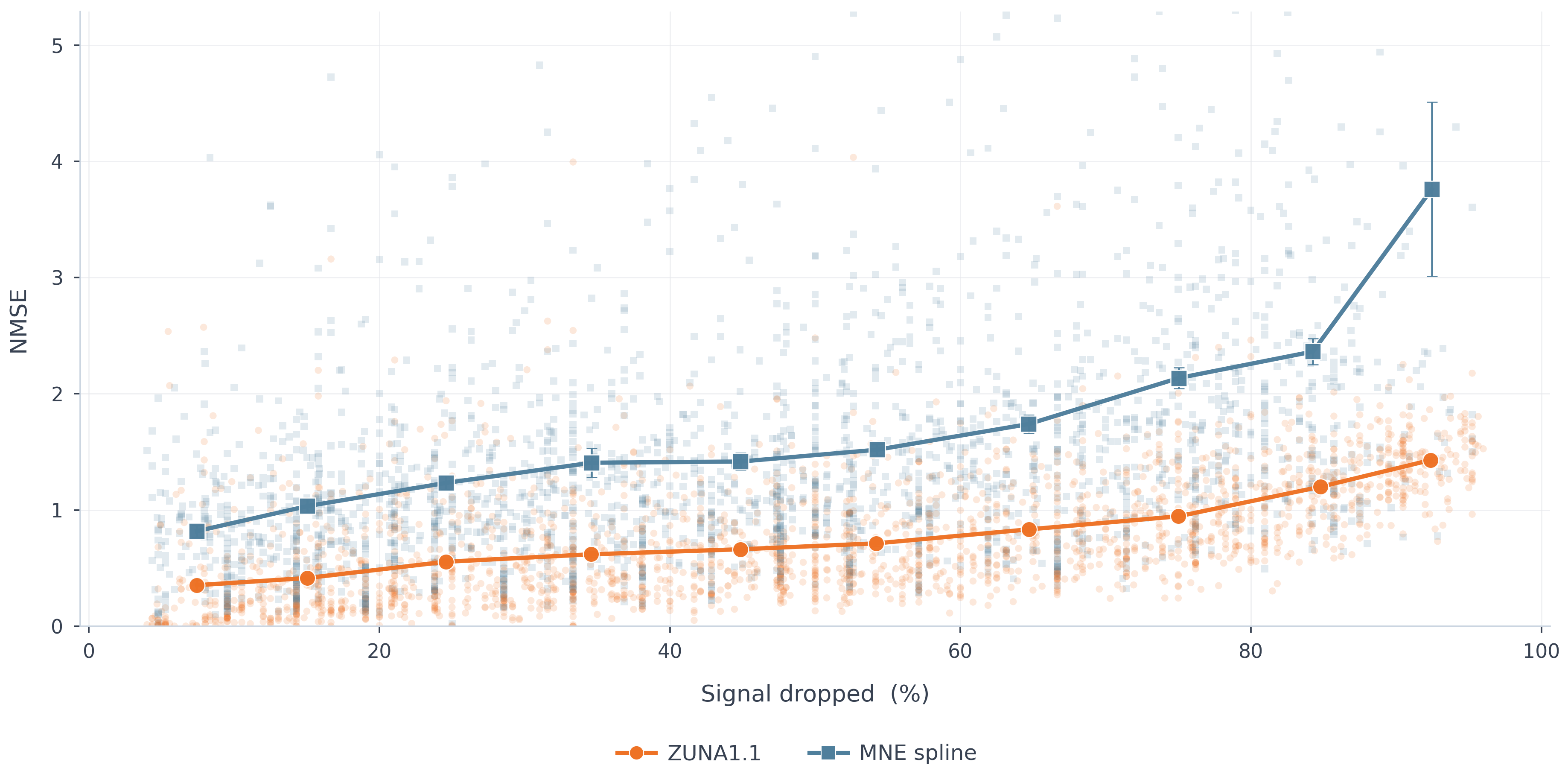}
        \caption{Full Channel Random Dropout}
        \label{fig:sub1}
    \end{subfigure}
    \hfill
    \begin{subfigure}[b]{0.32\textwidth}
        \centering
        \includegraphics[width=\textwidth]{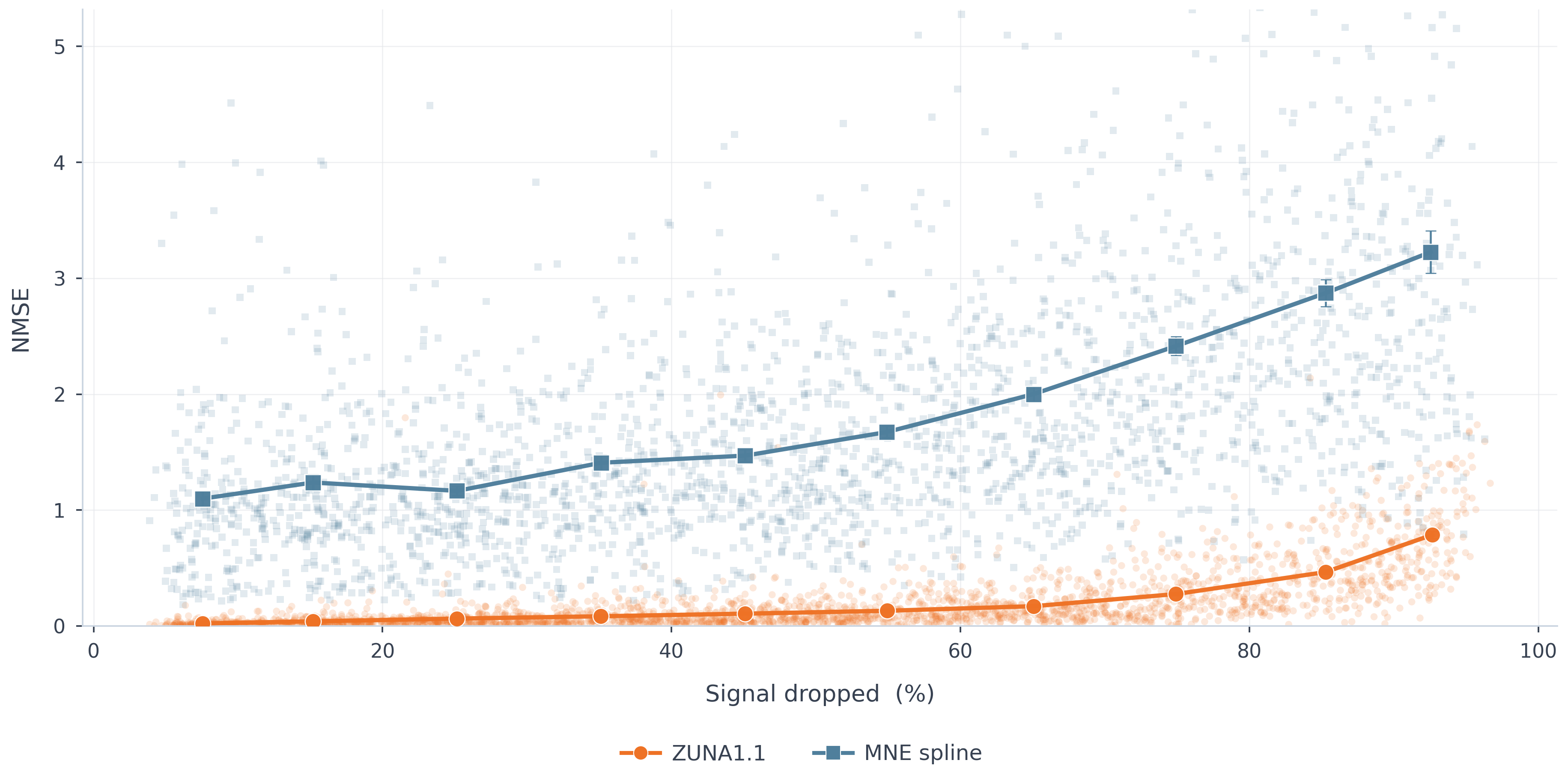}
        \caption{Random-Uniform Dropout}
        \label{fig:sub2}
    \end{subfigure}
    \hfill
    \begin{subfigure}[b]{0.32\textwidth}
        \centering
        \includegraphics[width=\textwidth]{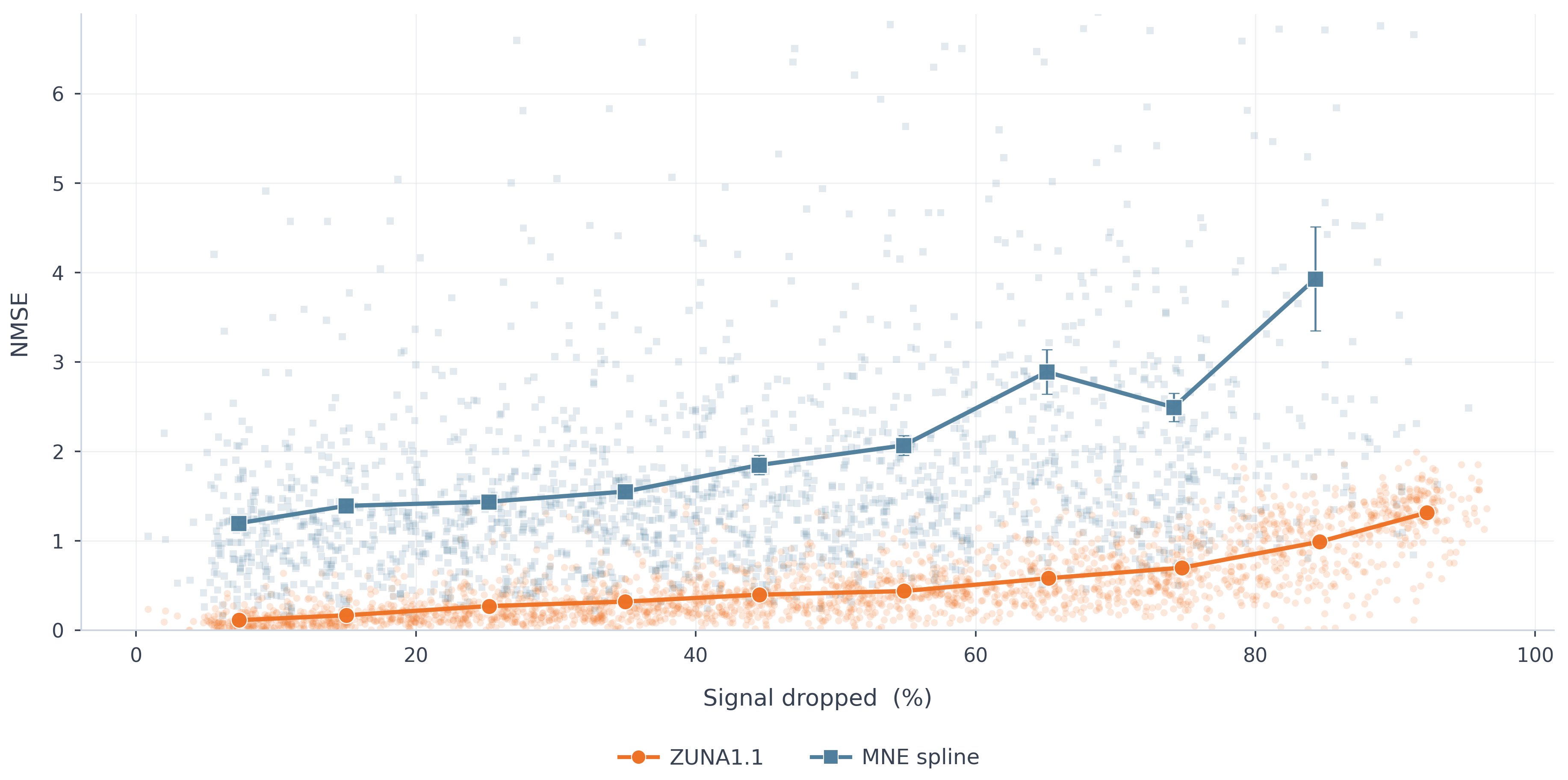}
        \caption{Spatio-Temporally Correlated Dropout}
        \label{fig:sub3}
    \end{subfigure}
    \caption{\textbf{\texttt{ZUNA1.1} vs MNE Spherical Spline} - reconstruction performance vs dropout percent on mixed evaluation dataset for structural dropout schemes}
    \label{fig:dropout_ZUNA_v_spline}
\end{figure*}

Finally, one can apply the spherical spline interpolation method to impute missing data in three of the four structural dropout schemes, using only information from observed channels at a given time-point to predict missing channels at that same time-point.  \texttt{ZUNA1.1} outperforms spline interpolation in these use-cases as well, as seen in Figure \ref{fig:dropout_ZUNA_v_spline}.  \texttt{ZUNA1.1} can use observed information from surrounding channels at the same time-point as well as observed signals from the same channel at previous or later time-points if signal degrades in a channel or group of channels for a short period of time.

\section{Discussion}

\texttt{ZUNA1.1} builds upon and substantially extends our original \texttt{ZUNA1} model, making it far more flexible and ultimately usable in real-world situations by EEG scientists and practitioners. The new dropout training schemes we introduced into the training of \texttt{ZUNA1.1} significantly expand its range and ability to process and reconstruct realistic ``in-the-wild'' EEG data from a variety of sources. The ability of \texttt{ZUNA1.1} to handle variable length sequences (up to 30 seconds) as well as reconstruct arbitrary time windows within a channel are crucial here. Beyond the dropout schemes, we also substantially improve the size and quality of our training dataset, and train for $4\times$ longer on $75\%$ more data than the original \texttt{ZUNA1} model. Architecture and training recipe improvements enabled us to train stably far beyond the original model, while computational efficiency improvements meant that training required only roughly the same level of computational resources. 

While \texttt{ZUNA1.1} substantially improves over its predecessor in its range and versatility, on the specific tasks which are suitable for \texttt{ZUNA1}, the two models are roughly comparable, with \texttt{ZUNA1.1} achieving only a small advantage on average. Both models still substantially outperform non-neural reconstruction methods such as MNE's spherical spline method. Since \texttt{ZUNA1.1} is trained for substantially longer on a much greater quantity (and higher quality) of data, this is somewhat surprising. One possibility is that the \texttt{ZUNA1.1} model is generally superior but that due to having to handle a much wider range of data, the model has to spread its competence over a much wider distribution, and hence necessarily dilute its capabilities in the specific kind of task which the original \texttt{ZUNA1} model has specialized in. This implies that the model is primarily limited by capacity. However, preliminary experiments with naively scaling up the model architecture from 380M to approximately 1.2B parameters did not show dramatically improved performance versus the smaller model, which led to \texttt{ZUNA1.1} being trained at the same size as the original \texttt{ZUNA1} model. This implies that parameter count alone is not the primary bottleneck for the model's capabilities.

Additional possibilities are that the diffusion encoder-decoder architecture of \texttt{ZUNA1.1} is itself the bottleneck. It is possible that the architectural scheme of passing the core information through the encoder bottleneck in `registers' imposes a heavy constraint on the ultimate reconstruction capabilities of the model, although expanding this bottleneck naively does not immediately result in better reconstructions. The encoder-decoder architecture itself could be a bottleneck. Another possibility is that the diffusion training scheme and especially the chosen noise schedule reduced the potential SNR of training causing an effective plateau in performance. 

Finally, it is of course possible, although we think it is unlikely, that even at 380M parameters, the model has already exhausted all of the signal present in the data which could be used for reconstruction. If this were true, it would mean that the irreducible noise floor in the EEG data itself is relatively high while the information redundancy across channels and time is fairly low. Some evidence that indirectly supports this is that we observed a similarly unexpected phenomenon with our initial set of dropout schemes, where a few of them appeared extremely challenging to learn such that the model made effectively little progress over the course of training. These were the intuitively harder dropout schemes such as the consumer or standard montage channel dropout. The lack of improvement on these schemes seems to imply that good reconstructions in these cases may be largely impossible in that the dropout scheme irreversibly removes the information required to make good predictions. 

A further point is that during development of \texttt{ZUNA1.1}, we observed a growing dissociation between reconstructive accuracy and the utility of the learned representations for downstream tasks. We discuss these explorations in appendix \cref{app:downstream}. While uncertain, we hypothesize that a substantial fraction of this dissociation may be due to the diffusion objective and architecture, which in other modalities is known to produce less clearly generalizable representations than more standard autoregressive objectives \citep{he2021maskedautoencoder}. More generally, for the goal of constructing powerful representations for downstream classifiers, pure reconstruction may not be the optimal objective in general, and latent-space objectives such as contrastive \citep{oord2018representation, he2019momentum, chen2020simple, radford2021learning} and JEPA-like \citep{lecun2022path, assran2023self} losses may be preferred. In future work, we aim to focus more on developing general encoder models of EEGs with representations well-suited to downstream tasks. This is ultimately a necessary step to establish the latent space which can then be joined with pretrained language or image models to train effective EEG-to-X models. This will likely require moving beyond regularized reconstruction as an objective, and developing novel architectures to move beyond \texttt{ZUNA1.1}'s diffusion autoencoder base. 

\section*{Acknowledgements}

We would like to thank Paul White, Danny Martinelli, and Kristina Zhao for their assistance with the release, and Nathan Kolbas for integrating \texttt{ZUNA1.1} into Zyphra Cloud.

%\newpage
%\clearpage

\bibliographystyle{tmlr}
\bibliography{main}

\clearpage
\appendices

\section{Training hyperparameters}
\label{app:hparams}

\begin{table}[h]
\centering
\caption{\textbf{ZUNA1.1 training configuration.}}
\label{tab:hparams}
\begin{tabular}{ll}
\toprule
Hyperparameter & Value \\
\midrule
Model dimension & 1024 \\
Layers (encoder / decoder) & 16 / 16 \\
Head dimension & 64 \\
Total parameters & 380M \\
Token size & 32 samples ($0.125$\,s at $256$\,Hz) \\
RoPE & 4D over $(x, y, z, t)$, $\theta = 10^4$ \\
Spatial discretization & 100 bins per axis \\
Max coarse-time steps & 240 ($30$\,s) \\
\midrule
Optimizer & AdamW \\
$\beta_1$, $\beta_2$ & $0.9$, $0.95$ \\
Weight decay & $0.1$ \\
LR schedule & cosine, warmup, min $= 0.01 \times$ peak \\
Peak learning rate & 5e-4 \\
Optimizer steps & 580k \\
Precision & \texttt{bf16} (norms, residual in \texttt{fp32}) \\
GPUs & 48 (6 nodes $\times$ 8) \\
Gradient accumulation & 2 \\
Packed tokens per rank & $22{,}000$ \\
Global batch & $\approx 2$M tokens / step \\
Encoder MMD weight & $10^{-2}$ $\longrightarrow$ 0\\
\midrule
Sample rate & $256$\,Hz \\
Window lengths & $0.5$--$30$\,s (\cref{sec:method:varlen}) \\
Quality thresholds (any / mean) & $0.1$ / $0.3$ \\
Filter variants & notch, bandpass (\cref{sec:method:filters}) \\
Dropout probability $p_\text{drop}$ & $0.9$ $\longrightarrow$ $0.99$ \\
Dropout mixture & 8 schemes, uniform (\cref{sec:method:dropout}) \\
\midrule
Inference ODE solver & Euler, 50 steps \\
Initial noise scale & $0.1$ \\
\bottomrule
\end{tabular}
\end{table}

%\newpage
\clearpage

% \clearpage

\section{Reconstruction performance}
\label{app:recon_tables}

\begin{table}[H]
\centering
\caption{Average NMSE for region occlusion reconstruction across datasets, methods, and target regions.}
\label{tab:region_occlusion_nmse}
\begin{tabular}{llccc}
\toprule
Dataset & Region & ZUNA1.1 & ZUNA1 & Spherical-spline \\
\midrule
ANPHY-Sleep & Frontal L & 0.241 $\pm$ 0.019 & \textbf{0.207 $\pm$ 0.016} & 0.547 $\pm$ 0.035 \\
 & Frontal R & \textbf{0.210 $\pm$ 0.011} & 0.234 $\pm$ 0.021 & 0.453 $\pm$ 0.049 \\
 & Temporal L & \textbf{0.393 $\pm$ 0.021} & 0.465 $\pm$ 0.036 & 0.718 $\pm$ 0.078 \\
 & Temporal R & \textbf{0.368 $\pm$ 0.015} & 0.449 $\pm$ 0.035 & 0.800 $\pm$ 0.087 \\
 & Central & 0.100 $\pm$ 0.009 & \textbf{0.069 $\pm$ 0.005} & 0.169 $\pm$ 0.018 \\
 & Parietal & 0.111 $\pm$ 0.012 & 0.111 $\pm$ 0.007 & \textbf{0.105 $\pm$ 0.011} \\
 & Occipital L & \textbf{0.091 $\pm$ 0.007} & 0.136 $\pm$ 0.010 & 0.184 $\pm$ 0.027 \\
 & Occipital R & \textbf{0.079 $\pm$ 0.007} & 0.119 $\pm$ 0.008 & 0.212 $\pm$ 0.023 \\
\midrule
BerlinBCI & Frontal L & \textbf{0.625 $\pm$ 0.048} & 0.650 $\pm$ 0.050 & 1.709 $\pm$ 0.185 \\
 & Frontal R & \textbf{0.608 $\pm$ 0.047} & 0.634 $\pm$ 0.049 & 1.615 $\pm$ 0.175 \\
 & Temporal L & \textbf{0.511 $\pm$ 0.035} & 0.565 $\pm$ 0.024 & 1.399 $\pm$ 0.151 \\
 & Temporal R & \textbf{0.506 $\pm$ 0.049} & 0.566 $\pm$ 0.051 & 1.334 $\pm$ 0.144 \\
 & Central & \textbf{0.328 $\pm$ 0.015} & 0.335 $\pm$ 0.026 & 0.642 $\pm$ 0.080 \\
 & Parietal & 0.405 $\pm$ 0.031 & \textbf{0.353 $\pm$ 0.027} & 0.513 $\pm$ 0.056 \\
 & Occipital L & \textbf{0.228 $\pm$ 0.018} & 0.248 $\pm$ 0.019 & 0.521 $\pm$ 0.021 \\
 & Occipital R & 0.194 $\pm$ 0.015 & \textbf{0.188 $\pm$ 0.014} & 0.404 $\pm$ 0.044 \\
\midrule
BCI2000 & Frontal L & \textbf{0.240 $\pm$ 0.028} & 0.289 $\pm$ 0.032 & 0.400 $\pm$ 0.053 \\
 & Frontal R & \textbf{0.249 $\pm$ 0.019} & 0.360 $\pm$ 0.028 & 0.410 $\pm$ 0.064 \\
 & Temporal L & \textbf{0.404 $\pm$ 0.041} & 0.468 $\pm$ 0.036 & 0.714 $\pm$ 0.067 \\
 & Temporal R & \textbf{0.400 $\pm$ 0.027} & 0.449 $\pm$ 0.035 & 0.637 $\pm$ 0.069 \\
 & Central & \textbf{0.216 $\pm$ 0.007} & 0.223 $\pm$ 0.017 & 0.328 $\pm$ 0.044 \\
 & Parietal & \textbf{0.164 $\pm$ 0.013} & 0.185 $\pm$ 0.014 & 0.183 $\pm$ 0.030 \\
 & Occipital L & \textbf{0.132 $\pm$ 0.019} & 0.165 $\pm$ 0.013 & 0.196 $\pm$ 0.021 \\
 & Occipital R & \textbf{0.098 $\pm$ 0.008} & 0.141 $\pm$ 0.009 & 0.138 $\pm$ 0.011 \\
\midrule
AAD & Frontal L & \textbf{0.798 $\pm$ 0.062} & 0.802 $\pm$ 0.062 & 1.233 $\pm$ 0.134 \\
 & Frontal R & 0.330 $\pm$ 0.006 & \textbf{0.327 $\pm$ 0.045} & 0.753 $\pm$ 0.082 \\
 & Temporal L & \textbf{0.998 $\pm$ 0.077} & 1.033 $\pm$ 0.079 & 1.030 $\pm$ 0.072 \\
 & Temporal R & \textbf{0.767 $\pm$ 0.059} & 0.903 $\pm$ 0.069 & 1.311 $\pm$ 0.142 \\
 & Central & \textbf{0.311 $\pm$ 0.024} & 0.528 $\pm$ 0.041 & 1.656 $\pm$ 0.179 \\
 & Parietal & 0.994 $\pm$ 0.047 & 1.014 $\pm$ 0.068 & \textbf{0.814 $\pm$ 0.088} \\
 & Occipital L & \textbf{1.001 $\pm$ 0.087} & 1.009 $\pm$ 0.057 & 1.064 $\pm$ 0.115 \\
 & Occipital R & \textbf{1.004 $\pm$ 0.058} & 1.026 $\pm$ 0.089 & 1.017 $\pm$ 0.089 \\
\bottomrule
\end{tabular}
\end{table}

\begin{table}[H]
      \centering
      \caption{Channel reconstruction NMSE on dropped channels. Lower is better.}
      \label{tab:reconstruction-nmse-fig4-replot}
      \begin{tabular}{llccc}
      \toprule
      Dataset & Dropout (\%) & ZUNA1.1 & ZUNA1 & Spherical-spline \\
      \midrule
      ANPHY-Sleep & 20 & 0.342 $\pm$ 0.015 & 0.422 $\pm$ 0.128 & \textbf{0.328 $\pm$ 0.024} \\
       & 50 & 0.419 $\pm$ 0.071 & \textbf{0.395 $\pm$ 0.050} & 0.496 $\pm$ 0.076 \\
       & 75 & \textbf{0.475 $\pm$ 0.043} & 0.633 $\pm$ 0.090 & 0.637 $\pm$ 0.035 \\
       & 90 & \textbf{0.745 $\pm$ 0.088} & 0.845 $\pm$ 0.067 & 1.443 $\pm$ 0.156 \\
      BerlinBCI & 20 & 0.602 $\pm$ 0.039 & \textbf{0.573 $\pm$ 0.018} & 0.762 $\pm$ 0.027 \\
       & 50 & 0.737 $\pm$ 0.022 & \textbf{0.674 $\pm$ 0.017} & 1.040 $\pm$ 0.033 \\
       & 75 & 1.039 $\pm$ 0.052 & \textbf{0.840 $\pm$ 0.018} & 1.616 $\pm$ 0.092 \\
       & 90 & 1.486 $\pm$ 0.069 & \textbf{1.135 $\pm$ 0.025} & 2.718 $\pm$ 0.583 \\
      BCI2000 & 20 & \textbf{0.367 $\pm$ 0.014} & 0.391 $\pm$ 0.021 & 0.385 $\pm$ 0.021 \\
       & 50 & \textbf{0.419 $\pm$ 0.020} & 0.429 $\pm$ 0.020 & 0.457 $\pm$ 0.023 \\
       & 75 & \textbf{0.507 $\pm$ 0.027} & 0.573 $\pm$ 0.021 & 0.777 $\pm$ 0.040 \\
       & 90 & \textbf{0.711 $\pm$ 0.020} & 0.931 $\pm$ 0.024 & 1.715 $\pm$ 0.159 \\
      AAD & 20 & 0.890 $\pm$ 0.001 & \textbf{0.764 $\pm$ 0.001} & 0.970 $\pm$ 0.001 \\
       & 50 & 0.845 $\pm$ 0.001 & \textbf{0.796 $\pm$ 0.001} & 0.898 $\pm$ 0.001 \\
       & 75 & 0.831 $\pm$ 0.001 & \textbf{0.804 $\pm$ 0.001} & 1.061 $\pm$ 0.001 \\
       & 90 & \textbf{0.836 $\pm$ 0.001} & 0.846 $\pm$ 0.001 & 1.608 $\pm$ 0.001 \\
      \bottomrule
      \end{tabular}
      \end{table}

\clearpage

\section{Downstream Tasks}
\label{app:downstream}

We evaluated \texttt{ZUNA1.1} on a range of downstream classification tasks. We probed the latent representation with a linear decoder that averages over tokens and finetunes the encoder as in NeuralBench \citep{Neuralbench2026}. We note that frozen backbones with linear heads performed significantly worse than finetuned models for \texttt{ZUNA1.1}, which is consistent across EEG foundation models (FMs) as found in NeuroAtlas \citep{kontras2026neuroatlas} and EEG-FM-Bench \citep{xiong2026eegfmbench}. 

However, we note that we find no correlation between model size and classification performance on finetuned FMs, not only in NeuralBench but also when scaling \texttt{ZUNA1.1} up to 1.2B parameters, which performed worse than the current 380M model. A simple linear regression across average model ranking in classification and number of parameters yielded $R^2=0.007$ with no obvious trends across eight FMs, including both \texttt{ZUNA1.1-380M} (i.e. \texttt{ZUNA1.1}) and \texttt{ZUNA1.1-1.2B}. Our experiments and the NeuralBench results suggest that new architectures and training objectives may be needed, and simply adding parameters is not a panacea to representation learning.

\begin{table}[H]
    \centering
    \caption{\textbf{Classification accuracy} (0-1 scale) for the MMD loss ablation across downstream tasks. Bold indicates the highest classification accuracies. We find a peak early on in our accuracies, and then a sharp decay as the latent representations are increasingly specialized towards reconstruction.}
    \label{tab:mmd_ablation_final}
    \small
    \setlength{\tabcolsep}{3pt}
    \resizebox{\columnwidth}{!}{%
    \begin{tabular}{lcccc}
    \toprule
    Dataset & No MMD-50K & MMD--50K & Hybrid--400K & Hybrid--575K \\
    \midrule
    Audiovisual
    & \textbf{0.53 $\pm$ 0.05}
    & 0.40 $\pm$ 0.03
    & 0.36 $\pm$ 0.04
    & 0.35 $\pm$ 0.03 \\

    CVEP
    & 0.88 $\pm$ 0.02
    & \textbf{0.88 $\pm$ 0.09}
    & 0.87 $\pm$ 0.01
    & 0.71 $\pm$ 0.07 \\

    Dementia
    & \textbf{0.49 $\pm$ 0.01}
    & 0.45 $\pm$ 0.01
    & 0.46 $\pm$ 0.01
    & 0.47 $\pm$ 0.02 \\

    Depression
    & 0.89 $\pm$ 0.01
    & \textbf{0.91 $\pm$ 0.01}
    & 0.89 $\pm$ 0.01
    & 0.90 $\pm$ 0.02 \\

    ERN
    & \textbf{0.81 $\pm$ 0.01}
    & 0.81 $\pm$ 0.03
    & 0.80 $\pm$ 0.02
    & 0.80 $\pm$ 0.02 \\

    LRP
    & 0.77 $\pm$ 0.02
    & \textbf{0.78 $\pm$ 0.01}
    & 0.78 $\pm$ 0.01
    & 0.50 $\pm$ 0.01 \\

    Mental Arithmetic
    & \textbf{0.73 $\pm$ 0.02}
    & 0.68 $\pm$ 0.02
    & 0.67 $\pm$ 0.02
    & 0.62 $\pm$ 0.05 \\

    Mental Imagery
    & \textbf{0.26 $\pm$ 0.01}
    & 0.25 $\pm$ 0.01
    & 0.22 $\pm$ 0.01
    & 0.20 $\pm$ 0.01 \\

    N170
    & 0.72 $\pm$ 0.01
    & \textbf{0.72 $\pm$ 0.01}
    & 0.70 $\pm$ 0.01
    & 0.58 $\pm$ 0.06 \\

    N2PC
    & 0.58 $\pm$ 0.01
    & \textbf{0.62 $\pm$ 0.02}
    & 0.55 $\pm$ 0.02
    & 0.50 $\pm$ 0.01 \\

    N400
    & \textbf{0.63 $\pm$ 0.01}
    & 0.61 $\pm$ 0.02
    & 0.61 $\pm$ 0.02
    & 0.59 $\pm$ 0.04 \\
    \bottomrule
    \end{tabular}%
    }
  \end{table}

\begin{table}[H]
      \centering
      \caption{\textbf{Reconstruction NMSE} under 50\% random-uniform dropout. Bold indicates the lowest errors, showing that our final checkpoint is almost uniformly superior for reconstruction. }
      \label{tab:mmd_ablation_reconstruction}
      \small
      \setlength{\tabcolsep}{3pt}
      \resizebox{\columnwidth}{!}{%
      \begin{tabular}{lcccc}
      \toprule
      Dataset & No MMD-50K & MMD--50K & Hybrid--400K & Hybrid--575K \\
      \midrule
      Audiovisual
      & 0.53 $\pm$ 0.02
      & 0.53 $\pm$ 0.02
      & 0.51 $\pm$ 0.02
      & \textbf{0.50 $\pm$ 0.02} \\
      Mental Arithmetic
      & 1.01 $\pm$ 0.04
      & 1.05 $\pm$ 0.04
      & \textbf{1.00 $\pm$ 0.04}
      & 1.06 $\pm$ 0.04 \\
      Mental Imagery
      & 0.56 $\pm$ 0.04
      & 0.66 $\pm$ 0.04
      & 0.55 $\pm$ 0.04
      & \textbf{0.54 $\pm$ 0.04} \\
      CVEP
      & 1.22 $\pm$ 0.06
      & 1.88 $\pm$ 0.05
      & \textbf{1.19 $\pm$ 0.07}
      & 1.24 $\pm$ 0.00 \\
      Dementia
      & 0.77 $\pm$ 0.05
      & 0.81 $\pm$ 0.05
      & 0.77 $\pm$ 0.06
      & \textbf{0.76 $\pm$ 0.05} \\
      Depression
      & 0.83 $\pm$ 0.07
      & 0.86 $\pm$ 0.07
      & 0.83 $\pm$ 0.07
      & \textbf{0.82 $\pm$ 0.07} \\
      ERN
      & 1.16 $\pm$ 0.04
      & 1.08 $\pm$ 0.01
      & 1.03 $\pm$ 0.03
      & \textbf{1.02 $\pm$ 0.01} \\
      LRP
      & 1.09 $\pm$ 0.02
      & 1.09 $\pm$ 0.01
      & 1.05 $\pm$ 0.01
      & \textbf{1.02 $\pm$ 0.01} \\
      N170
      & 1.08 $\pm$ 0.02
      & 1.06 $\pm$ 0.01
      & \textbf{1.01 $\pm$ 0.01}
      & 1.02 $\pm$ 0.01 \\
      N400
      & 1.14 $\pm$ 0.04
      & 1.10 $\pm$ 0.01
      & \textbf{1.02 $\pm$ 0.01}
      & \textbf{1.02 $\pm$ 0.01} \\
      N2PC
      & 1.07 $\pm$ 0.02
      & 1.08 $\pm$ 0.01
      & 1.05 $\pm$ 0.02
      & \textbf{1.03 $\pm$ 0.01} \\
      \bottomrule
      \end{tabular}%
      }
    \end{table}

In particular, there seems to be a tension between masked autoencoders and downstream representations. \citet{elouahidi2025reve} argues that the final layer of the encoder in a masked autoencoder is prone to overfitting to the reconstruction task, harming representations for downstream adaptation. However, they provide no ablations for their regularization that attempts to mitigate this issue. We provide experiments to show how this tension manifests in classification and reconstruction evaluations. 

In Table \ref{tab:mmd_ablation_final} and Table \ref{tab:mmd_ablation_reconstruction}, we show the counterintuitive relationship between classification accuracy and reconstruction error: as the number of training steps increases, reconstruction accuracy \textit{increases} while classification accuracy \textit{decreases}. The training objective contains a term that rewards reconstruction (decoder RF-loss) and a term that encourages latent representations to be more Gaussian distributed (encoder MMD-loss); however, there is no explicit term that shapes latents to be informative or useful for downstream tasks. At first, in the style of REVE \citep{elouahidi2025reve}, we hypothesized that our loss function was not encouraging useful latent structure, and we ablated our Maximum Mean Discrepancy (MMD) loss that encourages latent Gaussian structure. Deleting MMD did seem to help classification accuracy by about 2\% per task on average, and motivated our switch to turning off MMD loss in our main training run at training step 400K (i.e. our ''Hybrid'' variation trained with MMD loss up to 400K training steps, and turned off thereafter). 

%However, the MMD loss ablation was nowhere near the level of difference in evaluations across the simple number of training steps.
With or without the MMD loss, classification accuracy plateaued early in training and then decreased sharply thereafter. Our final checkpoint was almost uniformly superior on reconstruction while becoming progressively worse at classification. Even after turning off the MMD loss, which proved to be useful in our first ablation, the classification accuracies degraded until some of the datasets were even near-chance on classification accuracy even though the model performed extremely well at reconstruction. 

\begin{figure*}[t]
    \centering
    \includegraphics[width=\linewidth]{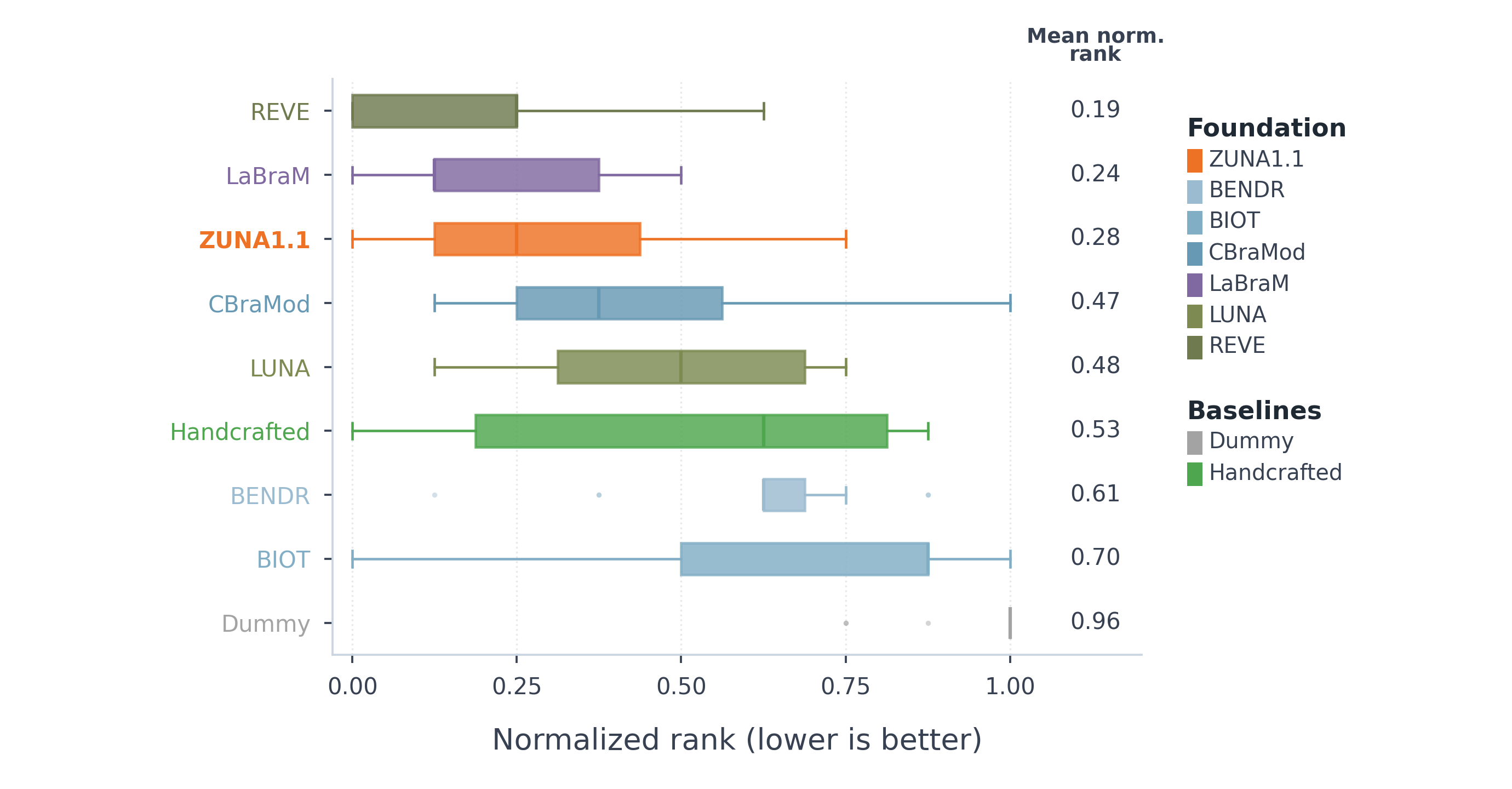}
    \caption{\textbf{Model ranking on downstream tasks. } Models are ordered by their mean normalized rank, displayed on the right hand side of the boxes. Lower is better. ZUNA1.1 is shown for No MMD-50K. }
    \label{fig:boxplot-ranking}
\end{figure*}

In Figure \ref{fig:boxplot-ranking}, we show the model rankings on EEG foundation models including \texttt{ZUNA1.1} using the NeuralBench metrics. Here, the ``No MMD-50K'' checkpoint yields respectable performance on the NeuralBench datasets. However, our results cast serious doubt on the usage of masked autoencoders trained to reconstruct missing data for downstream tasks. If REVE were trained longer with more data, would its classification accuracy also decrease given its objective is entirely reconstructive? Without these ablations, the answer is inconclusive. However, this seems to be the case for \texttt{ZUNA1.1}.  

\begin{figure*}[t]
      \centering
      \begin{subfigure}{0.49\textwidth}
          \centering
          \includegraphics[width=\linewidth]{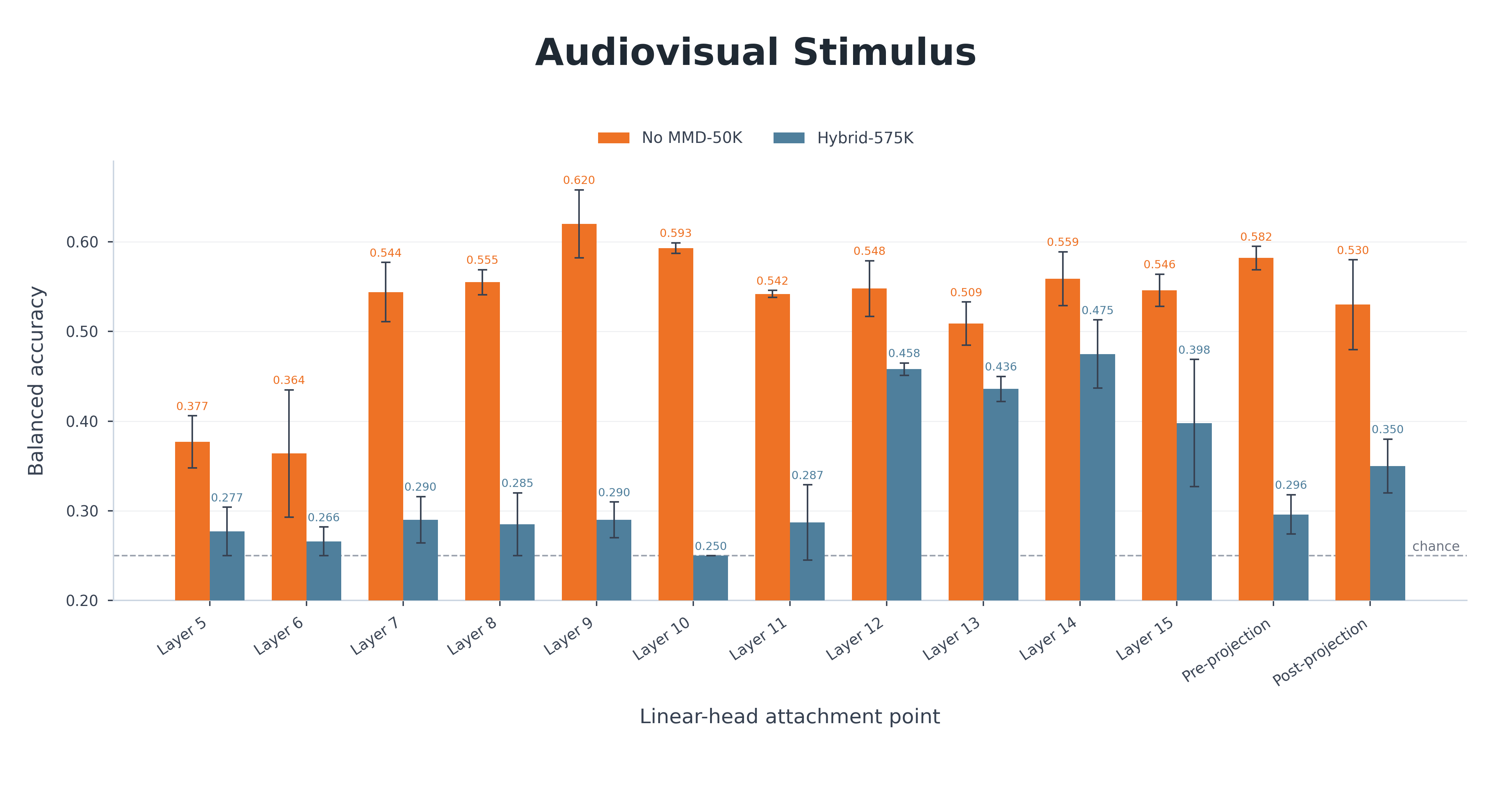}
          \caption{Audiovisual}
      \end{subfigure}
      \hfill
      \begin{subfigure}{0.49\textwidth}
          \centering
          \includegraphics[width=\linewidth]{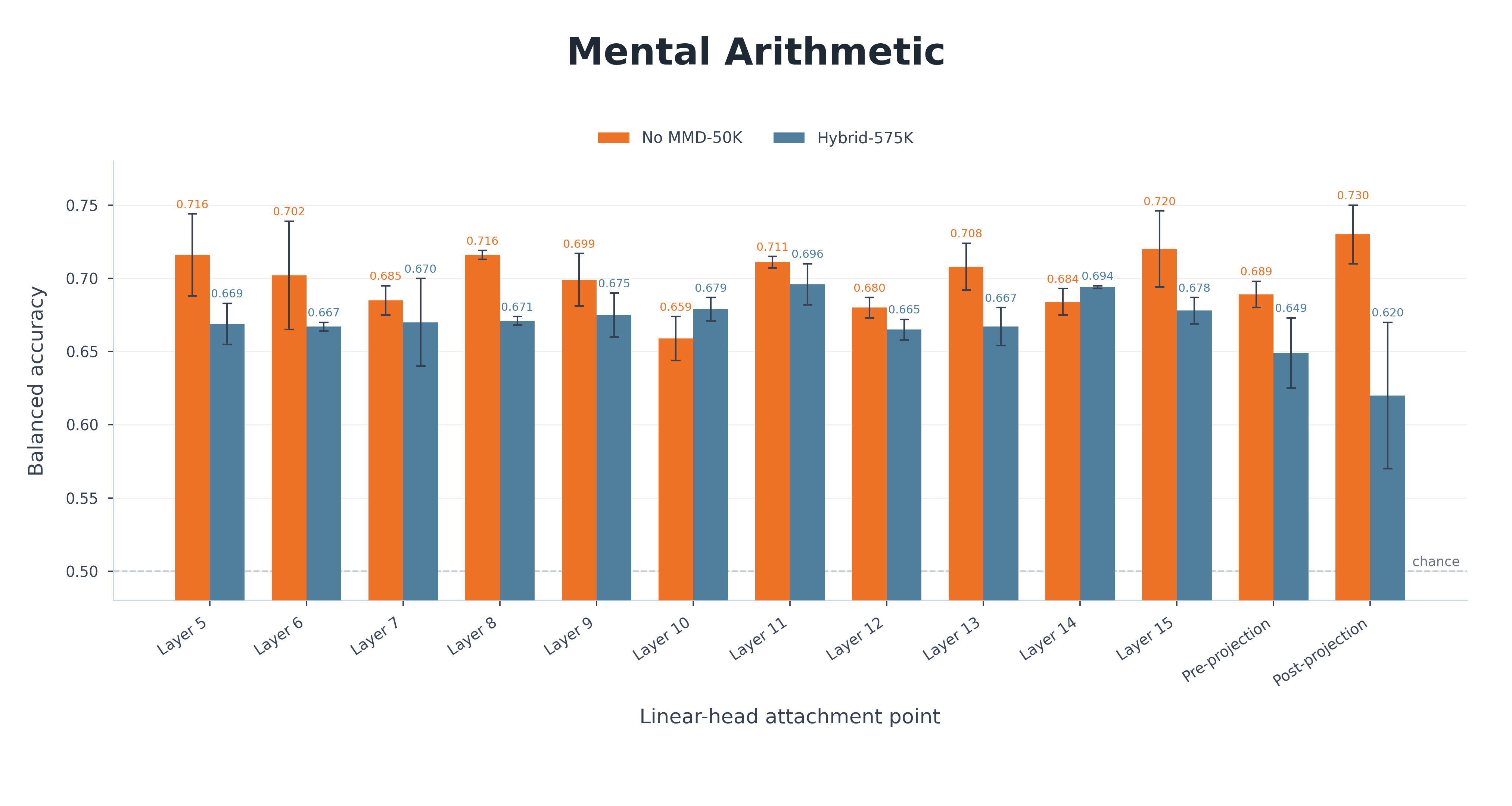}
          \caption{Mental Arithmetic}
      \end{subfigure}

      \vspace{0.5em}

      \begin{subfigure}{0.49\textwidth}
          \centering
          \includegraphics[width=\linewidth]{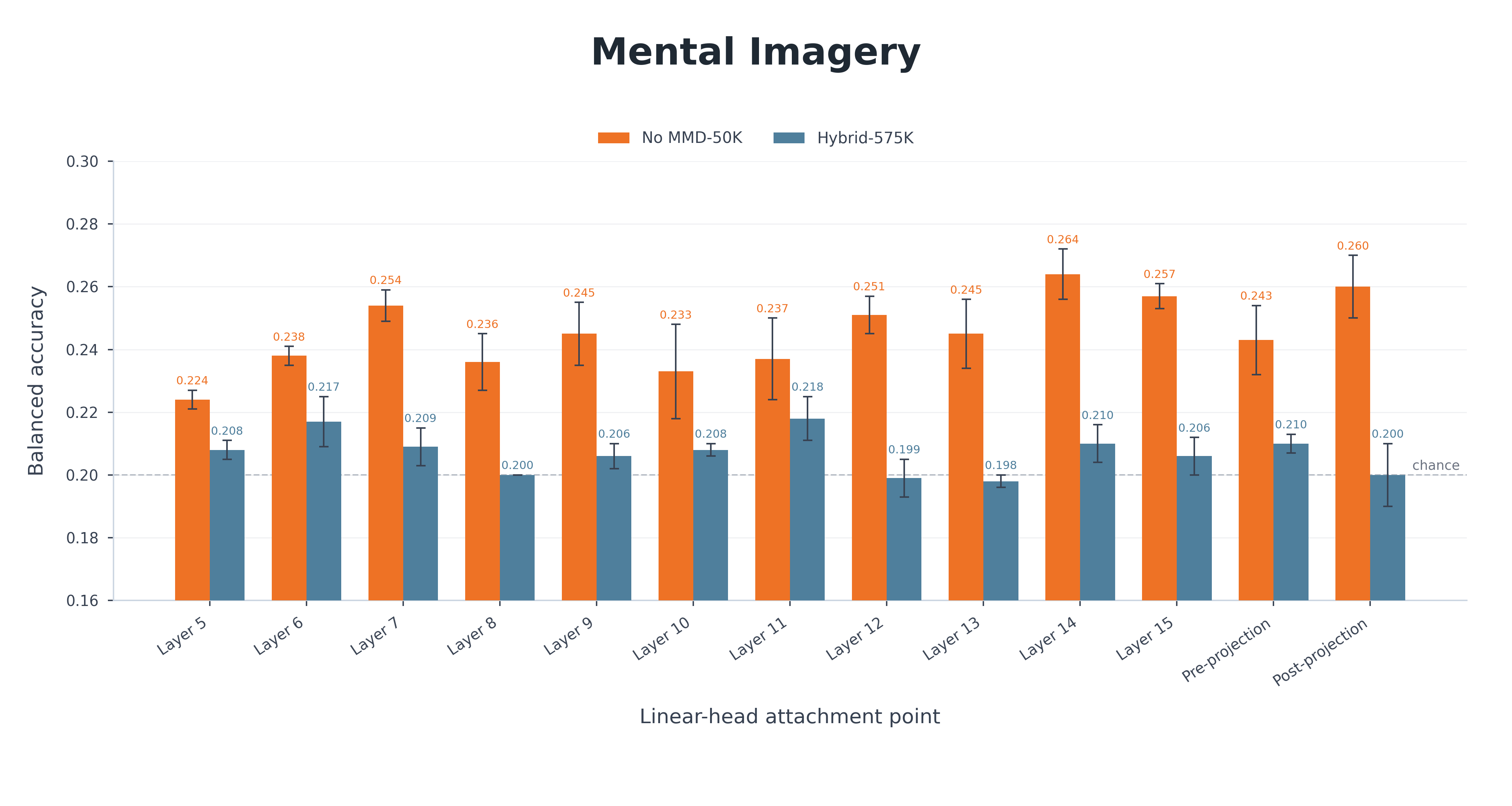}
          \caption{Mental Imagery}
      \end{subfigure}
      \hfill
      \begin{subfigure}{0.49\textwidth}
          \centering
          \includegraphics[width=\linewidth]{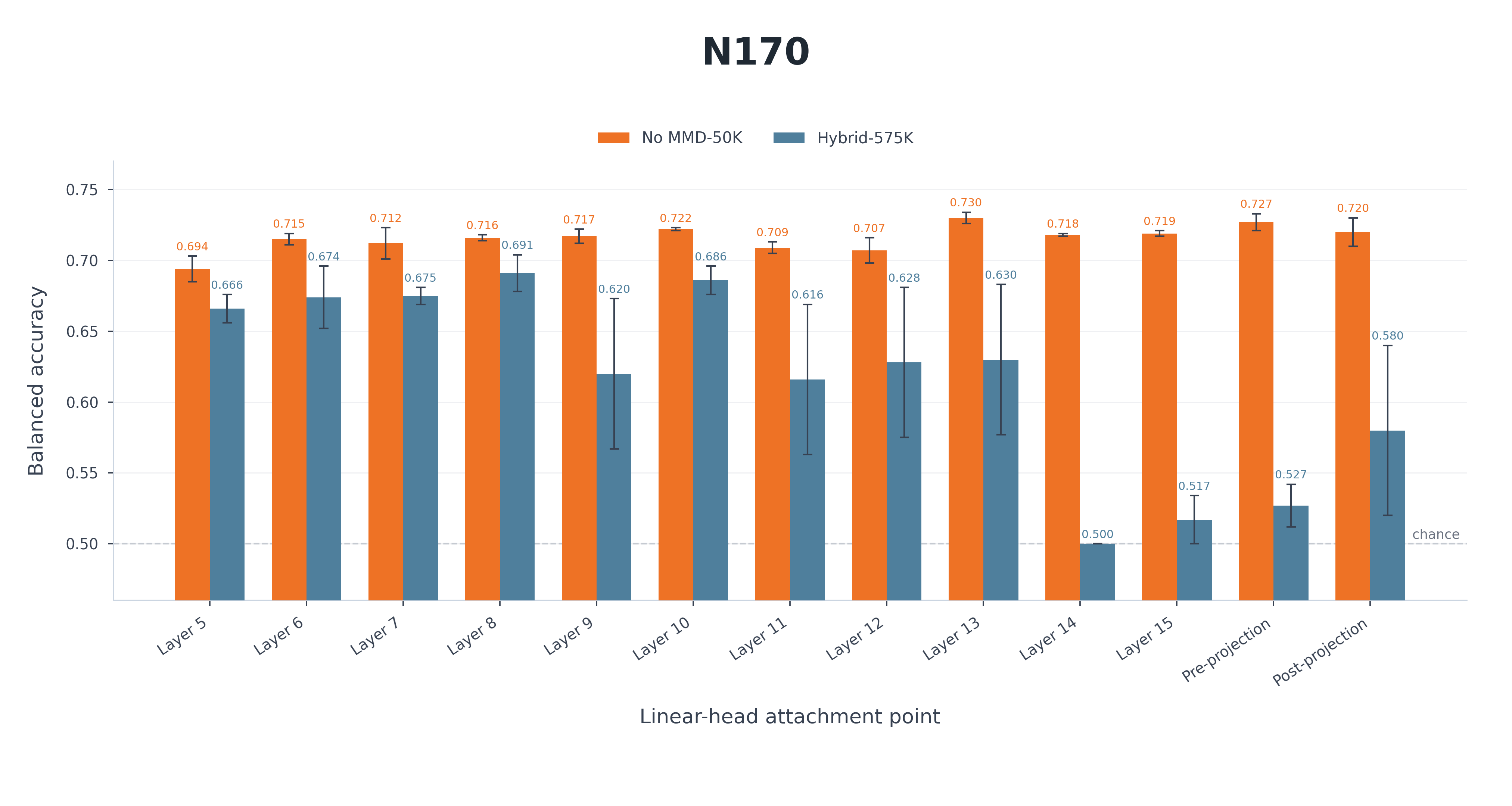}
          \caption{N170}
      \end{subfigure}

      \vspace{0.5em}

      \begin{subfigure}{0.49\textwidth}
          \centering
          \includegraphics[width=\linewidth]{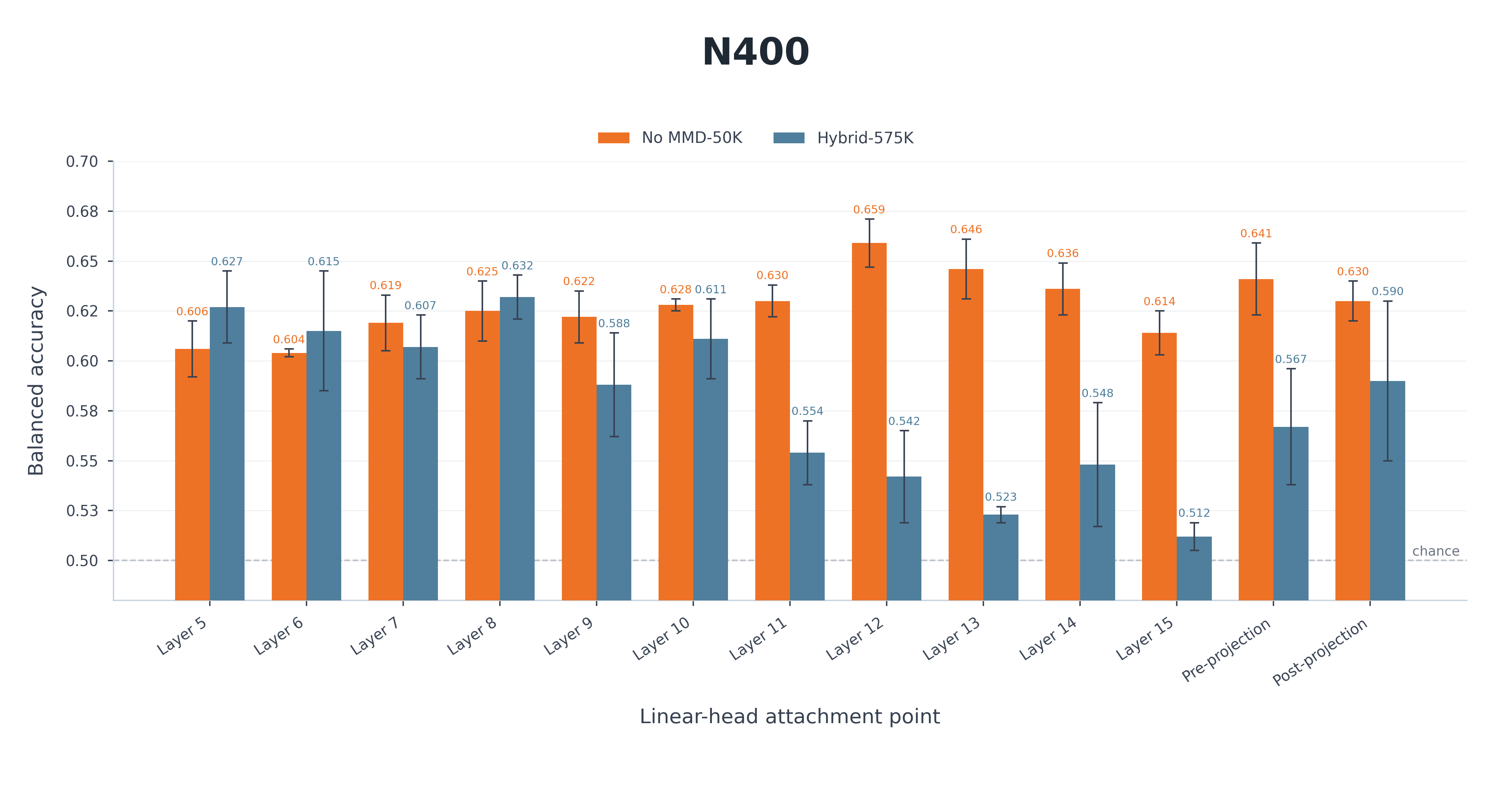}
          \caption{N400}
      \end{subfigure}

      \caption{Balanced accuracy for linear classifiers attached at different ZUNA
      representation layers. We ran this experiment to probe the representational quality due to previous hypotheses in EEG literature that the final layer of masked autoencoders is prone to overfitting to the reconstruction task. We do not see uniform evidence of this across datasets.}
      \label{fig:classifiers_by_layer}
  \end{figure*}

Finally, investigating the claim that the final encoder layer was overfit to the reconstruction task, we systematically explored whether building a linear classifier head on earlier layers of the encoder improves classification performance. Figure \ref{fig:classifiers_by_layer} shows balanced accuracy for classifiers built on top of different layers in the encoder for a few datasets in the NerualBench benchmark.  We observed some improved classification performance when using representations from earlier encoder layers, but nothing systematic.  We include these results for completeness.

  \begin{table*}[t]
  \centering
  \caption{Classification accuracy for ZUNA1.1 (No MMD) compared with NeuralBench EEG
  foundation models and baseline methods.}
  \label{tab:zuna11_neuralbench_comparison}
  \scriptsize
  \setlength{\tabcolsep}{3pt}
  \begin{tabular}{lccccccccc}
  \toprule
  Dataset & ZUNA1.1 & BENDR & BIOT & CBraMod & LaBraM & LUNA & REVE & Dummy & Handcrafted \\
  \midrule
  Artifact & 0.431$\pm$0.016 & 0.502$\pm$0.017 & 0.439$\pm$0.016 & 0.503$\pm$0.003 & 0.518$
  \pm$0.015 & 0.502$\pm$0.015 & \textbf{0.539$\pm$0.039} & 0.141$\pm$0.000 & 0.421$\pm$0.000
  \\
  Audiovisual & 0.528$\pm$0.054 & 0.419$\pm$0.020 & 0.261$\pm$0.023 & 0.311$\pm$0.076 &
  0.366$\pm$0.095 & 0.262$\pm$0.016 & 0.505$\pm$0.010 & 0.250$\pm$0.000 & \textbf{0.623$
  \pm$0.000} \\
  CVEP & 0.875$\pm$0.024 & 0.188$\pm$0.240 & 0.047$\pm$0.003 & 0.871$\pm$0.013 &
  \textbf{0.876$\pm$0.005} & 0.872$\pm$0.009 & 0.791$\pm$0.008 & 0.050$\pm$0.000 & 0.048$
  \pm$0.000 \\
  Dementia & \textbf{0.486$\pm$0.011} & 0.331$\pm$0.010 & 0.428$\pm$0.027 & 0.328$\pm$0.017 &
  0.413$\pm$0.023 & 0.443$\pm$0.049 & 0.408$\pm$0.045 & 0.333$\pm$0.000 & 0.475$\pm$0.001 \\
  Depression & \textbf{0.892$\pm$0.004} & 0.857$\pm$0.019 & 0.867$\pm$0.018 & 0.873$\pm$0.006
  & 0.867$\pm$0.038 & 0.844$\pm$0.003 & 0.870$\pm$0.006 & 0.500$\pm$0.000 & 0.813$\pm$0.001 \\
  ERN & 0.814$\pm$0.013 & 0.768$\pm$0.046 & 0.464$\pm$0.036 & 0.811$\pm$0.015 &
  \textbf{0.833$\pm$0.008} & 0.819$\pm$0.011 & 0.814$\pm$0.033 & 0.500$\pm$0.001 & 0.775$
  \pm$0.000 \\
  LRP & 0.774$\pm$0.016 & 0.705$\pm$0.004 & 0.502$\pm$0.004 & 0.780$\pm$0.007 & 0.789$
  \pm$0.010 & 0.747$\pm$0.006 & \textbf{0.792$\pm$0.006} & 0.500$\pm$0.001 & 0.652$\pm$0.001
  \\
  Mental Arithmetic & 0.732$\pm$0.021 & 0.681$\pm$0.012 & 0.640$\pm$0.023 & 0.685$\pm$0.021 &
  0.687$\pm$0.075 & 0.643$\pm$0.069 & \textbf{0.739$\pm$0.027} & 0.500$\pm$0.000 & 0.698$
  \pm$0.000 \\
  Mental imagery & 0.261$\pm$0.012 & 0.209$\pm$0.011 & \textbf{0.275$\pm$0.017} & 0.202$
  \pm$0.003 & 0.218$\pm$0.008 & 0.209$\pm$0.014 & 0.256$\pm$0.021 & 0.200$\pm$0.000 & 0.263$
  \pm$0.000 \\
  Motor execution & 0.498$\pm$0.010 & 0.338$\pm$0.012 & 0.238$\pm$0.016 & 0.510$\pm$0.004 &
  0.539$\pm$0.010 & 0.489$\pm$0.014 & \textbf{0.588$\pm$0.023} & 0.062$\pm$0.000 & 0.176$
  \pm$0.000 \\
  N170 & 0.715$\pm$0.006 & 0.662$\pm$0.009 & 0.504$\pm$0.005 & 0.707$\pm$0.008 & 0.728$
  \pm$0.003 & 0.693$\pm$0.006 & \textbf{0.755$\pm$0.013} & 0.500$\pm$0.000 & 0.663$\pm$0.000
  \\
  N2PC & 0.584$\pm$0.012 & 0.533$\pm$0.030 & 0.503$\pm$0.003 & 0.616$\pm$0.006 & 0.631$
  \pm$0.022 & 0.607$\pm$0.015 & \textbf{0.653$\pm$0.003} & 0.500$\pm$0.000 & 0.507$\pm$0.000
  \\
  N400 & 0.634$\pm$0.009 & 0.616$\pm$0.003 & 0.506$\pm$0.009 & 0.611$\pm$0.022 & 0.639$
  \pm$0.007 & 0.648$\pm$0.012 & 0.645$\pm$0.011 & 0.500$\pm$0.000 & \textbf{0.661$\pm$0.000}
  \\
  P300 & \textbf{0.670$\pm$0.028} & 0.665$\pm$0.002 & 0.502$\pm$0.001 & 0.630$\pm$0.001 &
  0.652$\pm$0.004 & 0.626$\pm$0.003 & 0.654$\pm$0.001 & 0.500$\pm$0.000 & 0.601$\pm$0.000 \\
  SSVEP & 0.231$\pm$0.072 & 0.113$\pm$0.071 & 0.589$\pm$0.029 & 0.490$\pm$0.247 &
  \textbf{0.965$\pm$0.002} & 0.118$\pm$0.004 & 0.961$\pm$0.002 & 0.025$\pm$0.000 & 0.391$
  \pm$0.000 \\
  \bottomrule
  \end{tabular}
  \end{table*}

\clearpage

\end{document}